%% file: main.tex
\documentclass{article} % For LaTeX2e
\usepackage[preprint]{colm2026_conference}

\usepackage{microtype}
\usepackage{hyperref}
\usepackage{url}
\usepackage{booktabs}
\usepackage{array,tabularx,longtable}
\newcolumntype{P}[1]{>{\raggedright\arraybackslash}p{#1}}
\usepackage{lineno}
\usepackage{amsmath,amsfonts,amssymb}
\usepackage{tikz}
\usetikzlibrary{trees, shapes.geometric, arrows.meta, positioning, calc, shadows, backgrounds, fit}
\usepackage[edges]{forest}
\usepackage{enumitem}
\usepackage{multirow}
\usepackage{graphicx}
\usepackage{placeins}
\usepackage{fontawesome}

\definecolor{deepblue}{HTML}{356580}
\definecolor{midblue}{HTML}{2980B9}
\definecolor{lightblue}{HTML}{AED6F1}
\definecolor{lightblue2}{HTML}{D4E6F1}
\definecolor{deeppurple}{HTML}{705885}
\definecolor{deepgray}{HTML}{242426}
\definecolor{midpurple}{HTML}{8E44AD}
\definecolor{lightpurple}{HTML}{D2B4DE}
\definecolor{deeporange}{HTML}{A45B35}
\definecolor{midorange}{HTML}{E67E22}
\definecolor{lightorange}{HTML}{F5CBA7}
\definecolor{coralpink}{HTML}{E74C3C}
\definecolor{midgray}{HTML}{65727C}
\definecolor{panelbg}{HTML}{F5F8FC}
\definecolor{darkblue}{rgb}{0, 0, 0.5}
\hypersetup{colorlinks=true, citecolor=darkblue, linkcolor=darkblue, urlcolor=darkblue}

\title{A Survey of On-Policy Distillation for Large Language Models}

\author{Mingyang Song, Mao Zheng\\
Foundation Model Department, Tencent, China\\
\texttt{nickmysong@tencent.com}\\[2pt]
\faGithub\ \href{https://github.com/nick7nlp/Awesome-LLM-On-Policy-Distillation}{nick7nlp/Awesome-LLM-On-Policy-Distillation}}

\newcommand{\loss}{\mathcal{L}}

\fancypagestyle{surveytitle}{%
  \fancyhead[L]{\raisebox{0pt}[0pt][0pt]{\includegraphics[height=17pt]{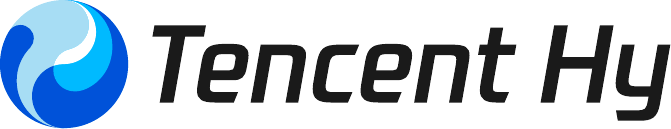}}}%
}

\begin{document}
\maketitle
\fancyhead[L]{}
\thispagestyle{surveytitle}
\raggedbottom

\begin{abstract}
On-Policy Distillation (OPD) trains large language models with teacher guidance on student-generated trajectories, addressing the mismatch between fixed demonstrations and the histories encountered during deployment. As its use expands across language model training, different formulations of supervision and optimization make it difficult to compare methods and understand when they are effective. This survey develops a unified view of OPD through the interplay between student-generated states, teacher information, and policy updates. This framework connects the foundations of distillation with self-distillation, multi-teacher integration, reinforcement learning, and interactive agents. We examine how these settings change the supervision available to the student and relate the resulting design choices to capability transfer, training efficiency, and deployment. Our synthesis centers on the conditions under which guidance remains reliable and learnable as the student evolves, considering both immediate performance and the capacity for subsequent learning. We conclude by discussing open challenges in adapting supervision and allocating computation to support effective learning.
\end{abstract}

\input{chapters/01_introduction}

\input{chapters/02_foundations}
\input{chapters/03_teacher}
\input{chapters/04_self_distillation}
\input{chapters/05_rl_opd}
\input{chapters/06_mopd}
\input{chapters/07_agentic}
\input{chapters/08_training}
\input{chapters/09_understanding}
\input{chapters/10_applications}
\input{chapters/11_open_problems}
\input{chapters/12_conclusion}

\clearpage
\bibliographystyle{colm2026_conference}
\bibliography{references,references_background}
\end{document}

%% file: chapters/01_introduction.tex
\section{Introduction}
\label{sec:intro}

Knowledge distillation trains a \emph{student} model using guidance from a \emph{teacher}. It can transfer a larger model's capabilities to a smaller one, combine several specialists in one model, or teach behavior that initially requires extra context. Across these uses, the student must continue from its own outputs, which can lead it away from the histories represented in its training examples. Fixed responses and teacher demonstrations provide no direct guidance at many of these unfamiliar histories~\citep{1606.07947,1011.0686}. Even close imitation of a teacher's answer style can leave a gap in underlying competence~\citep{2305.15717}.

\textbf{On-Policy Distillation (OPD) brings teacher guidance to the student's own generation process.} The student generates a response, and the teacher supplies guidance along it. For text generation, a \emph{state} consists of the prompt and the response prefix generated so far. A \emph{trajectory} is the resulting sequence of states and generated tokens. If the student takes a wrong turn, OPD can provide guidance at that prefix even when a fixed teacher answer would never visit it. MiniLLM and GKD develop complementary realizations through sequence-level optimization and matching next-token distributions~\citep{2306.08543,2306.13649}. Both connect student-generated experience to teacher guidance, whose value depends on the visited state and the student's ability to learn. Figure~\ref{fig:opd_loop} depicts this learning loop and the design choices examined in the following chapters.

A capable teacher can still be misleading after an erroneous student prefix, motivating target adaptation and reliability checks~\citep{2604.13016,2609.16459}. \textbf{Useful supervision depends on what the teacher knows at the student's current state.} Conversely, additional information can create a teaching advantage without a separate stronger model. On-policy self-distillation gives a version of the student a solution, demonstration, or feedback, then uses its predictions as supervision. This information is \emph{privileged} because the deployed student will not normally receive it. OPSD, SDFT, and SDPO use these information sources to construct supervision when a suitable stronger model is unavailable or impractical~\citep{2601.18734,2601.19897,2601.20802}.

The choice of supervisor is intertwined with the task objective. Teacher preferences guide individual predictions, while reinforcement learning (RL) can use outcome rewards to evaluate complete attempts. Joint updates, selective supervision, and successive training stages offer different ways to combine these signals~\citep{2506.02208,2604.02288,2609.04108}. With several teachers, training must also decide which specialist to consult and how to balance learning across capabilities. Individually effective specialists do not automatically produce a student that retains all their strengths~\citep{2606.30406,2608.19098}. These design choices can operate together. A self-teacher may interpret feedback from several specialists, and reward learning may be interleaved with transfer between evolving models~\citep{2609.02548,2604.27083}.

For interactive agents, the student's actions also change the environment in which later decisions are made. Tools return observations and memory changes the visible history, so the state must include more than the student's generated response prefix. Delayed feedback must inform earlier decisions, and teacher intervention can change the trajectory itself~\citep{2604.24005,2608.07068}. In both response generation and interaction, learning changes which states the student reaches next. Deciding what to collect, supervise, and reuse therefore affects both the information available for learning and its computational cost~\citep{2605.05940,2606.24143}. The consequences extend beyond the distillation loss to task behavior, retained capabilities, and the capacity for subsequent learning~\citep{2606.26091,2609.28145}.

We organize the survey around \emph{the states visited by the student, the information used to construct supervision, and the update that supervision induces}. These distinctions explain why similar training procedures can behave differently. The same divergence can be evaluated on different histories, the same model can teach differently with additional context, and the same token score can enter different gradient estimators. Together, these choices connect the mechanisms of OPD to their effects on learning, transfer, and deployment.

Figure~\ref{fig:taxonomy} provides a reading map organized around these dependencies. Section~\ref{sec:background} builds from classical distillation to student-state supervision, capacity constraints, and optimization before Section~\ref{sec:signal} examines teacher supervision. Sections~\ref{sec:self_distillation},~\ref{sec:rl_opd}, and~\ref{sec:mopd} develop self-distillation, coordination with reinforcement learning, and multi-teacher integration. Sections~\ref{sec:agentic} and~\ref{sec:dynamics} then examine interaction and the allocation and reuse of supervision as policies change. Section~\ref{sec:understanding} synthesizes the evidence, which informs the practical comparisons and open questions in Sections~\ref{sec:applications} and~\ref{sec:future}.

\input{figures/taxonomy}

%% file: figures/taxonomy.tex
\begin{figure}[!t]
\centering
\begin{tikzpicture}[
  x=1cm,y=1cm,
  head/.style={anchor=north west,align=left,inner xsep=0.12cm,inner ysep=0.10cm,
    rounded corners=1.5pt,font=\fontsize{9}{11}\selectfont\bfseries},
  entry/.style={anchor=north west,align=left,inner sep=0pt,
    font=\fontsize{8.5}{10.3}\selectfont}
]
\definecolor{figureink}{HTML}{292C2E}
\definecolor{figuremuted}{HTML}{596263}
\definecolor{figureline}{HTML}{CDD2D0}
\definecolor{deeporange}{HTML}{945C43}
\definecolor{deeppurple}{HTML}{6B5C78}
\definecolor{deepblue}{HTML}{3D6871}
\definecolor{midgray}{HTML}{636F72}
\definecolor{maporange}{HTML}{F4ECE6}
\definecolor{mappurple}{HTML}{F0ECF2}
\definecolor{mapblue}{HTML}{EAF1F1}
\definecolor{mapgray}{HTML}{F2F3F2}
\tikzset{text=figureink}
\defcitealias{2306.13649}{GKD}
\defcitealias{2402.03898}{DistiLLM}
\defcitealias{2503.07067}{DistiLLM-2}
\defcitealias{2606.01039}{OPD+}
\defcitealias{2605.07865}{vOPD}
\defcitealias{2306.08543}{MiniLLM}
\defcitealias{2609.16937}{$\gamma$OPD}
\defcitealias{2511.10643}{GAD}
\defcitealias{2604.18963}{Distillation Traps and Guards}
\defcitealias{2609.02998}{TGOPD}
\defcitealias{2602.12125}{G-OPD}
\defcitealias{2607.05394}{Direct-OPD}
\defcitealias{2607.26246}{W2S-OPD}
\defcitealias{2607.22334}{BPM}
\defcitealias{2606.06021}{OPRD}
\defcitealias{2601.18734}{OPSD}
\defcitealias{2601.19897}{SDFT}
\defcitealias{2602.12275}{OPCD}
\defcitealias{2601.20802}{SDPO}
\defcitealias{2605.18740}{Vision-OPD}
\defcitealias{2605.00642}{GUI-SD}
\defcitealias{2506.02208}{KDRL}
\defcitealias{2604.02288}{SRPO}
\defcitealias{2608.24696}{OPDVR}
\defcitealias{2609.04108}{OPD-then-RL}
\defcitealias{2609.28145}{RL Starts before RL}
\defcitealias{2604.27083}{CoPD}
\defcitealias{2606.30406}{MOPD}
\defcitealias{2609.02548}{MT-SDPO}
\defcitealias{2609.23697}{TrustMOPD}
\defcitealias{2608.19098}{Open-MOPD}
\defcitealias{2605.07725}{SOD}
\defcitealias{2608.07068}{MemOPD}
\defcitealias{2607.04763}{Prefix Replay}
\defcitealias{2605.17873}{HINT-SD}
\defcitealias{2605.19447}{SERL}
\defcitealias{2607.29078}{DASH-OPD}
\defcitealias{2604.24005}{TCOD}
\defcitealias{2607.05804}{TurnOPD}
\defcitealias{2609.14636}{STRIDE}
\defcitealias{2609.04172}{One-shot OPD}
\defcitealias{2604.14084}{TIP}
\defcitealias{2605.21606}{PW-OPSD}
\defcitealias{2602.15260}{FOPD}
\defcitealias{2604.13010}{Lightning-OPD}
\defcitealias{2606.24143}{AsyncOPD}
\defcitealias{2505.09388}{Qwen3}
\defcitealias{deepseekv4}{DeepSeek-V4}
\hypersetup{citecolor=figureink,linkcolor=figuremuted}
\hyphenpenalty=10000\exhyphenpenalty=10000
\node[head,fill=mapblue,text=deepblue,text width=13.64cm] (basehead) at (0,0) {\ref{sec:background}\quad Foundations and optimization};
\node[entry,text width=13.64cm,text=figuremuted,below=0.11cm of basehead.south west,anchor=north west,xshift=0.09cm] (basepath) {\ref{subsec:classical_kd} Classical KD\hfill$\rightarrow$\hfill\ref{subsec:states} Student states\hfill$\rightarrow$\hfill\ref{subsec:formulation} Matching\hfill$\rightarrow$\hfill\ref{subsec:distillation_scaling} Capacity and scaling};
\node[entry,text width=4.30cm,below=0.16cm of basepath.south west,anchor=north west] (base0) {\textbf{\ref{subsec:support} Matching objectives\\ and support}\\ \mbox{\citetalias{2306.13649}}; \mbox{\citetalias{2402.03898}}; \mbox{\citetalias{2503.07067}}};
\node[entry,text width=4.30cm,below=0.16cm of basepath.south west,anchor=north west,xshift=4.69cm] (base1) {\textbf{\ref{subsec:estimators} Gradient estimation\\ and variance}\\ \mbox{\citetalias{2606.01039}}; \mbox{\citetalias{2605.07865}}};
\node[entry,text width=4.30cm,below=0.16cm of basepath.south west,anchor=north west,xshift=9.38cm] (base2) {\textbf{\ref{subsec:temporal_credit} Local and temporal credit}\\ \mbox{\citetalias{2306.08543}}; \mbox{\citetalias{2609.16937}}};
\node[fit=(base0)(base1)(base2),inner sep=0pt] (basefit) {};
\coordinate (col0) at ([yshift=-0.27cm,xshift=-0.09cm]basefit.south west);
\begin{scope}[shift={($(col0)+(0,0)$)}]
\node[head,fill=mappurple,text=deeppurple,text width=6.48cm] (teachhead) at (0,0) {\ref{sec:signal}\quad Teacher supervision};
\node[entry,text width=6.54cm,below=0.12cm of teachhead.south west,anchor=north west,xshift=0.09cm] (teachbody) {\textbf{\ref{subsec:teacher_roles} Teacher roles and access}\\ \mbox{\citetalias{2306.13649}}; \mbox{\citetalias{2511.10643}}
\par\vspace{2.7pt}
\textbf{\ref{subsec:teacher_validity} Reliability and teacher adaptation}\\ \mbox{\citetalias{2604.18963}}; \mbox{\citetalias{2609.02998}}
\par\vspace{2.7pt}
\textbf{\ref{subsec:target_construction} Transformed targets and reference policies}\\ \mbox{\citetalias{2602.12125}}; \mbox{\citetalias{2607.05394}}; \mbox{\citetalias{2607.26246}}
\par\vspace{2.7pt}
\textbf{\ref{subsec:interfaces} Prediction and representation interfaces}\\ \mbox{\citetalias{2607.22334}}; \mbox{\citetalias{2606.06021}}};
\end{scope}
\begin{scope}[shift={($(teachbody.south west)+(-0.09,-0.27)$)}]
\node[head,fill=mappurple,text=deeppurple,text width=6.48cm] (selfhead) at (0,0) {\ref{sec:self_distillation}\quad On-policy self-distillation};
\node[entry,text width=6.54cm,below=0.12cm of selfhead.south west,anchor=north west,xshift=0.09cm] (selfbody) {\textbf{\ref{subsec:context_construction} Solutions, demonstrations, and experience}\\ \mbox{\citetalias{2601.18734}}; \mbox{\citetalias{2601.19897}}; \mbox{\citetalias{2602.12275}}
\par\vspace{2.7pt}
\textbf{\ref{subsec:self_feedback} Feedback-conditioned supervision}\\ \mbox{\citetalias{2601.20802}}
\par\vspace{2.7pt}
\textbf{\ref{subsec:privileged_views} Privileged perceptual views}\\ \mbox{\citetalias{2605.18740}}; \mbox{\citetalias{2605.00642}}
\par\vspace{2.7pt}
\textbf{\ref{subsec:self_evolution} Teacher evolution and regularization}\\ \mbox{\citetalias{2601.20802}}};
\end{scope}
\begin{scope}[shift={($(selfbody.south west)+(-0.09,-0.27)$)}]
\node[head,fill=mapblue,text=deepblue,text width=6.48cm] (rlhead) at (0,0) {\ref{sec:rl_opd}\quad OPD with reinforcement learning};
\node[entry,text width=6.54cm,below=0.12cm of rlhead.south west,anchor=north west,xshift=0.09cm] (rlbody) {\textbf{\ref{subsec:joint_training} Joint optimization}\\ \mbox{\citetalias{2306.13649}}; \mbox{\citetalias{2506.02208}}
\par\vspace{2.7pt}
\textbf{\ref{subsec:selective_training} Selective supervision}\\ \mbox{\citetalias{2604.02288}}; \mbox{\citetalias{2608.24696}}
\par\vspace{2.7pt}
\textbf{\ref{subsec:sequential_training} Sequential and alternating learning}\\ \mbox{\citetalias{2609.04108}}; \mbox{\citetalias{2609.28145}}; \mbox{\citetalias{2604.27083}}};
\end{scope}
\coordinate (col1) at ([yshift=-0.27cm,xshift=7.07cm]basefit.south west);
\begin{scope}[shift={($(col1)+(0,0)$)}]
\node[head,fill=mappurple,text=deeppurple,text width=6.48cm] (multihead) at (0,0) {\ref{sec:mopd}\quad Multi-teacher OPD (MOPD)};
\node[entry,text width=6.54cm,below=0.12cm of multihead.south west,anchor=north west,xshift=0.09cm] (multibody) {\textbf{\ref{subsec:teacher_composition} Teacher routing, eligibility,\\ and composition}\\ \mbox{\citetalias{2606.30406}}; \mbox{\citetalias{2609.02548}}; \mbox{\citetalias{2609.23697}}
\par\vspace{2.7pt}
\textbf{\ref{subsec:mopd_allocation} Capability balance; \ref{subsec:mopd_evolution} evolving specialists}\\ \mbox{\citetalias{2608.19098}}; \mbox{\citetalias{2604.27083}}};
\end{scope}
\begin{scope}[shift={($(multibody.south west)+(-0.09,-0.27)$)}]
\node[head,fill=mapgray,text=midgray,text width=6.48cm] (agenthead) at (0,0) {\ref{sec:agentic}\quad Agentic OPD};
\node[entry,text width=6.54cm,below=0.12cm of agenthead.south west,anchor=north west,xshift=0.09cm] (agentbody) {\textbf{\ref{subsec:interaction_units} Action tokens; \ref{subsec:state_alignment} memory and replay}\\ \mbox{\citetalias{2605.07725}}; \mbox{\citetalias{2608.07068}}; \mbox{\citetalias{2607.04763}}
\par\vspace{2.7pt}
\textbf{\ref{subsec:turn_credit} Cross-turn credit; \ref{subsec:intervention} teacher intervention}\\ \mbox{\citetalias{2605.17873}}; \mbox{\citetalias{2605.19447}}; \mbox{\citetalias{2607.29078}}
\par\vspace{2.7pt}
\textbf{\ref{subsec:agent_horizon} Interaction depth and restart}\\ \mbox{\citetalias{2604.24005}}; \mbox{\citetalias{2607.05804}}; \mbox{\citetalias{2609.14636}}};
\end{scope}
\begin{scope}[shift={($(agentbody.south west)+(-0.09,-0.27)$)}]
\node[head,fill=maporange,text=deeporange,text width=6.48cm] (trainhead) at (0,0) {\ref{sec:dynamics}\quad Training dynamics and efficiency};
\node[entry,text width=6.54cm,below=0.12cm of trainhead.south west,anchor=north west,xshift=0.09cm] (trainbody) {\textbf{\ref{subsec:coverage} Prompt breadth and state coverage}\\ \mbox{\citetalias{2609.04172}}
\par\vspace{2.7pt}
\textbf{\ref{subsec:selection} Token selection, weighting,\\ and generation horizon}\\ \mbox{\citetalias{2604.14084}}; \mbox{\citetalias{2605.21606}}; \mbox{\citetalias{2602.15260}}
\par\vspace{2.7pt}
\textbf{\ref{subsec:freshness} Refresh, caching, and policy lag}\\ \mbox{\citetalias{2608.19098}}; \mbox{\citetalias{2604.13010}}; \mbox{\citetalias{2606.24143}}
\par\vspace{2.7pt}
\textbf{\ref{subsec:lifecycle} Initialization; \ref{subsec:retention} retention}\\ \mbox{\citetalias{2505.09388}}; \mbox{\citetalias{2601.19897}}
\par\vspace{2.7pt}
\textbf{\ref{subsec:systems} Systems and teacher-scoring cost}\\ \mbox{\citetalias{deepseekv4}}};
\end{scope}
\end{tikzpicture}
\caption{Chapter map and representative studies for OPD. Shared foundations connect the design choices in Sections~\ref{sec:signal}--\ref{sec:dynamics}; methods may address several questions across groups. Section~\ref{sec:understanding} synthesizes the evidence, and Section~\ref{sec:applications} discusses practical use.}
\label{fig:taxonomy}
\end{figure}

%% file: chapters/02_foundations.tex
\section{OPD Foundations and Optimization}
\label{sec:background}
\label{sec:objectives}

Distillation transfers a teacher's predictions to a student, but autoregressive generation makes the context of that transfer part of the learning problem. Each prediction depends on earlier decisions, and the student can reach prefixes absent from the teacher's demonstrations. Understanding OPD therefore requires specifying both what the student should learn from the teacher and where that supervision should be applied. This distinction connects classical distillation to the objectives, resource choices, and optimization methods developed in this chapter.

\subsection{From Classical KD to Autoregressive Distillation}
\label{subsec:classical_kd}

In supervised learning, a one-hot target identifies the observed label. Classical knowledge distillation (KD) instead trains the student to match a teacher's predictive distribution, which also expresses relative preferences among alternative labels~\citep{1503.02531}. A temperature can soften these distributions, and the distillation loss can be combined with supervision from the original labels. These soft targets convey more information than a single predicted label, but transfer still depends on student capacity.

For an autoregressive language model, the corresponding prediction is a distribution over the next token. Let $x$ be an input, $y$ a response, and $c_t=(x,y_{<t})$ the prefix at position $t$. Given teacher probabilities $q(a\mid c_t)$ and student probabilities $\pi_\theta(a\mid c_t)$ over a shared vocabulary $V$, soft-target cross-entropy takes the form
\begin{equation}
\begin{aligned}
\ell_{\mathrm{KD}}(c_t)
 &=-\sum_{a\in V}q(a\mid c_t)\log\pi_\theta(a\mid c_t)\\
 &=H\!\left(q(\cdot\mid c_t)\right)
   +D_{\mathrm{KL}}\!\left(q(\cdot\mid c_t)\parallel\pi_\theta(\cdot\mid c_t)\right).
\end{aligned}
\label{eq:token_kd}
\end{equation}
With the teacher distribution held fixed, its entropy $H(q)$ is constant with respect to the student, so minimizing this loss is equivalent to minimizing forward Kullback--Leibler (KL) divergence. By comparison, supervised fine-tuning (SFT) predicts the supplied response token at each position. Sequence-level KD uses a teacher-generated response as that target sequence, allowing transfer through text without access to the teacher's full token distribution~\citep{1606.07947}. The broader LLM distillation survey of \citet{2402.13116} discusses these and other forms of transfer beyond OPD.

These objectives specify what to predict at a prefix, but leave open where that prefix comes from. When training follows reference or teacher-generated responses, the student learns under histories supplied by another source. At generation time, it conditions on its own earlier predictions. A loss that fits the supplied histories well can therefore leave the student poorly supervised at the histories it actually visits.

\subsection{The Student-State Learning Loop}
\label{subsec:scope}
\label{subsec:states}

\textbf{OPD brings teacher supervision to prefixes generated by the student.} Consider a student producing a mathematical solution. After an incorrect intermediate step, a teacher's original solution no longer describes how to continue from the student's history. OPD instead queries the teacher at that history, allowing supervision to respond to the student's own decisions. MiniLLM and GKD both incorporate student-generated sequences into distillation, while emphasizing sequence-level reverse KL and flexible conditional matching, respectively~\citep{2306.08543,2306.13649}. Changing the source of prefixes does not require abandoning the soft targets of classical KD.

The motivation predates language-model distillation. DAgger addresses the distribution shift between expert demonstrations and learner-visited states~\citep{1011.0686}. Under a bounded-cost assumption, a small imitation error on expert states can produce a worst-case cumulative cost gap of order $O(\epsilon T^2)$ over horizon $T$. Querying the expert on learner-visited states improves the horizon dependence when the expert can recover from learner mistakes and the learner controls its imitation error on those states. For language models, this motivates supervising student prefixes while leaving teacher reliability and the student's ability to learn from that guidance as substantive questions.

Strict on-policy collection draws supervised states from the current student. Reference responses, teacher responses, and responses from earlier students are off-policy relative to that student. This distinction concerns the source of training states, whereas \emph{offline} describes learning from a fixed collection. Mixture sampling changes the state source, and periodic collection or asynchronous execution can introduce policy lag~\citep{2306.13649,2402.03898,2605.05940}. An older student's trajectory was on-policy when collected but becomes stale during further optimization. Re-scoring it refreshes supervision without refreshing its state distribution. Similarly, a fresh student action after a replayed teacher history does not make that history student-generated~\citep{2607.04763}.

The learning signal matters as well as the source of trajectories. SPIN repeatedly contrasts responses from the previous model with fixed demonstrations, using sequence-level preferences rather than teacher distributions at student prefixes~\citep{2401.01335}. Iterative generation alone therefore does not identify the distillation loop. Figure~\ref{fig:opd_loop} separates collection from target construction and the subsequent student update.

\input{figures/opd_loop}

\subsection{A Common Framework for Distribution Matching}
\label{subsec:formulation}
\label{subsec:matching}

The prefix comparison above extends to other sequential decisions. Let $s$ denote a state and $\pi_\theta(a\mid s)$ the student policy. In autoregressive generation, $s=c_t$ and the action is the next token. An agent state additionally reflects observations and the memory visible to the model, while a diffusion-language state contains a partially decoded sequence and its denoising stage~\citep{2608.07068,2607.04428,2607.16872}. A teacher supplies a target $q(a\mid s^T)$, possibly using a view $s^T$ with information unavailable to the student. Teacher and student predictions must still concern comparable decisions.

A common matching objective is
\begin{equation}
\loss_{\mathrm{match}}(\theta)=\mathbb{E}_{s\sim d_\mu}
 \left[w(s)D_f\!\left(q(\cdot\mid s^T)\parallel\pi_\theta(\cdot\mid s)\right)\right],
\label{eq:common_match}
\end{equation}
where $d_\mu$ is the state distribution induced by a collection policy $\mu$, $w(s)$ is a nonnegative state weight, and $D_f$ measures disagreement between the target and student distributions. Equation~\ref{eq:token_kd} corresponds to forward-KL matching at a given prefix. Averaging over reference prefixes or current-student prefixes changes $d_\mu$ while retaining that conditional loss. Giving the teacher extra context instead changes $s^T$ and potentially $q$, even when the student sees the same state. These distinctions let us compare where supervision is applied, what information it conveys, and how disagreement is penalized.

For a convex generator $f$ with $f(1)=0$, the $f$-divergence is
\begin{equation*}
D_f(P\parallel Q)=\mathbb{E}_{a\sim Q}\left[f\!\left(\frac{P(a)}{Q(a)}\right)\right].
\end{equation*}
The \emph{support} of a distribution is the set of actions to which it assigns positive probability. Where the ratios are defined, $P=q$, $Q=\pi_\theta$ and $f(u)=u\log u$ give forward KL, while $f(u)=-\log u$ gives reverse KL~\citep{2307.15190}. Forward KL weights teacher-supported alternatives, whereas reverse KL penalizes student probability mass on actions the teacher considers unlikely. When the student cannot reproduce the target exactly, these weightings favor different approximations. Figure~\ref{fig:kl_divergence} illustrates the difference with a restricted student family, holding the state source fixed.

\input{figures/kl_comparison}

GKD makes the distinction between collection and matching explicit by varying both the fraction of student-generated responses and the conditional divergence~\citep{2306.13649}. MiniLLM introduces a further choice through its sequence-level reverse-KL objective, which accounts for how an action changes later prefixes~\citep{2306.08543}. Equation~\ref{eq:common_match} describes conditional matching, but does not determine the gradient estimator or whether to differentiate through the state distribution. We return to those choices after considering the capacity and resources available for transfer.

\subsection{Capacity and Distillation Scaling}
\label{subsec:distillation_scaling}

A matching objective specifies the desired transfer, but the student must learn it within a finite budget. A larger teacher can yield a weaker student when their capabilities differ substantially~\citep{2402.11890}. Distillation scaling studies make this constraint quantitative by relating student performance to model size, training data, and teacher quality. In pretraining-style distillation on English C4 reference prefixes, \citet{2502.08606} fit student validation cross-entropy as a function of student non-embedding parameters $N_S$, distillation tokens $D_S$, and teacher validation cross-entropy $L_T$. Within the studied model family, $L_T$ summarizes the influence of teacher size and pretraining tokens. The predicted outcome is cross-entropy on held-out data, rather than divergence from the teacher.

This relationship explains why choosing the strongest teacher need not yield the best student. At fixed student size and token budget, stronger teachers initially help, but further improvements in teacher quality can worsen student performance. Increasing the student's training budget changes which teacher transfers best, so the capacity gap involves learning resources as well as parameter counts~\citep{2502.08606}. A fixed compute budget adds another tradeoff. Scoring with a larger teacher consumes resources that could support more student training, while the cost of preparing that teacher can be shared across multiple uses. Teacher choice consequently depends on both predictive quality and the budget available for student learning.

These empirical scaling relationships concern pretraining distillation on reference data. In OPD, the distribution $d_\mu$ in Equation~\ref{eq:common_match} also changes as the student learns. Repeated generation from a fixed prompt can expose new prefixes, and replay can spend more updates on states already collected. The resource question consequently extends from how many tokens to use to which states should be generated, scored, and revisited. Section~\ref{sec:dynamics} examines these choices, while Section~\ref{subsec:future_scaling} considers what remains unresolved in their joint allocation. At each selected state, the form of supervision also determines which teacher--student discrepancies the student is asked to correct.

\subsection{Matching Objectives and Probability Support}
\label{subsec:support}

\textbf{An objective determines both how strongly discrepancies are penalized and which alternatives receive correction.} The $f$-divergence formulation makes the weighting explicit, but the probabilities available to the learner also matter. Full-distribution matching, sampled-token feedback, and restricted candidate sets expose different amounts of the teacher's preferences. Two questions therefore organize objective design: how much probability should move, and where should it go?

Consider first an update that scores only the student's sampled token. For autoregressive generation, let $p_\theta$ and $p_T$ denote the student and teacher sequence distributions, with $p_\theta(a\mid c_t)=\pi_\theta(a\mid c_t)$. A common sampled-token surrogate for conditional reverse-KL matching is
\begin{equation*}
 \ell_t(\theta)=-\mathrm{sg}[r_t]\log p_\theta(y_t\mid c_t),
 \qquad
 r_t=\log\frac{p_T(y_t\mid c_t)}{p_\theta(y_t\mid c_t)} ,
\end{equation*}
where $\mathrm{sg}$ treats its argument as constant during differentiation~\citep{2605.06387}. The coefficient $r_t$ is positive when the teacher assigns the sampled token more probability than the student, and negative in the opposite case. It therefore determines whether the update reinforces or suppresses that token.

Suppressing an action disfavored by the teacher does not identify a useful replacement. For a gradient step on the output \emph{logits}, the scores converted into probabilities by softmax, a detached negative coefficient increases each unchosen token's logit in proportion to its current student probability~\citep{2605.06387}. A teacher-preferred alternative with negligible student probability consequently receives little direct correction. This statement concerns logit changes; softmax normalization also affects the resulting probabilities. Full-vocabulary matching exposes all alternatives at a visited state, while top-$K$ supervision exposes a restricted set. Discarding omitted mass, renormalizing retained probabilities, or pooling the tail into a residual bucket yields different comparisons. These choices change the \emph{effective support of supervision}, not only its cost.

Additional forward guidance can expose destinations missed by sampled-token feedback, either selectively or at every position. AOPD supplies teacher top-$K$ alternatives when the sampled-token signal is non-positive; TGPO supplies the teacher's highest-probability token at every rollout position and anneals that correction during reward optimization~\citep{2605.06387,2605.13230}. Selecting where to intervene introduces a further choice. Proposed criteria include teacher entropy, the sampled teacher--student probability gap, and a clipped teacher-to-student probability ratio~\citep{2603.07079,2605.06387,2606.01249}. TrOPD uses the clipped ratio as the probability of routing a sampled token to reverse-KL supervision, with top-$K$ forward guidance otherwise. It also anneals teacher-generated prefixes before student continuations, coupling local routing with trajectory guidance. Its ``trust region'' measures predictive agreement, not teacher correctness.

The magnitude of correction matters when teacher and student probabilities differ sharply. Skew divergences moderate extreme ratios through probability mixtures; DistiLLM uses a fixed skew coefficient, while DistiLLM-2 adapts the skew and loss balance and applies forward matching to teacher responses and reverse matching to student responses~\citep{2402.03898,2503.07067}. The latter also reuses student responses sampled at the start of each epoch. A different simplification replaces the sampled log-ratio coefficient with its sign, scaling the shared gradient by the previous step's smoothed conditional-TV estimate~\citep{2609.08341}. This TV-OPD estimator concerns conditional total variation at fixed states; it does not assign credit for actions' effects on later states.

\textbf{Moderating correction and restoring missing support can pull in different directions.} Hellinger matching with geometric-mean weighting attenuates correction for teacher-preferred tokens that the student finds unlikely. GEOSD combines this weighting with a squared Fisher--Rao penalty against a periodically refreshed student reference~\citep{2607.06855}. Attenuation can limit extreme updates while also weakening guidance toward useful, low-probability continuations. Explicit mass-routing objectives address the destination separately: TIDE combines Hellinger-shaped suppression of excess mass with teacher top-$K$ cross-entropy to restore deficits, whereas RouteOPD couples excess and deficit tokens through log-odds changes fitted to a shared bounded teacher potential~\citep{2608.09836,2609.08337}. RouteOPD's controls favor teacher-deficit destinations over matched-rank or random alternatives. Stability of the correction alone therefore does not settle where it should be directed.

Relative-logit matching expresses the discrepancy through pairwise preferences. Concrete Score Distillation (CSD) compares weighted teacher--student logit differences across vocabulary pairs, making the comparison invariant to a common logit shift~\citep{2509.25837}. Factorized detached pair weights permit an $O(|V|)$ gradient, although a shared vocabulary is still required. The weights remain consequential: in its small-data GSM8K setting, student-only weighting fails while mixed teacher--student weighting remains effective. Across these objectives, the form of a discrepancy, its weighting, and the alternatives exposed to the student jointly determine the intended correction. The next question is how that correction is estimated.

\subsection{Estimating Updates and Controlling Variance}
\label{subsec:estimators}

The same named loss can lead to different updates, depending on what is sampled and what is differentiated. \textbf{A conditional update fits predictions at a fixed prefix; a sequence update also accounts for how actions change later prefixes.} The sampled-token surrogate in Section~\ref{subsec:support} holds both the prefix and the coefficient $r_t$ fixed during differentiation. Differentiating through $r_t$ would produce a different gradient, even on the same sampled token.

For teacher-to-student density ratio $u=p_T(y_t\mid c_t)/p_\theta(y_t\mid c_t)$, the general $f$-divergence coefficient $-f(u)+uf'(u)$ includes direct parameter dependence omitted by a detached $-f(u)$ surrogate~\citep{2606.01039}. Reverse KL is a special case because the added term is constant and its expected score contribution vanishes. The correction concerns the conditional gradient at a fixed prefix, leaving temporal credit as a separate choice.

The support choices in Section~\ref{subsec:support} also determine which source of sampling error an estimator can remove. A sampled log ratio needs only the teacher probability of the chosen token~\citep{lu2025onpolicy}, whereas full-vocabulary KL sums the conditional alternatives exactly, removing uncertainty from sampling the next action at that prefix. Top-$K$ supervision approximates or changes that comparison according to its tail treatment. Training still samples the prefixes themselves, and exact conditional summation does not automatically assign credit for future states. Access to more logits can therefore reduce conditional sampling error while leaving state freshness, trajectory variance, and task alignment unresolved.

Estimator studies separate centering a noisy signal from choosing which signals to emphasize. A \emph{baseline} subtracts a reference value from the sampled coefficient. If that value depends on the prefix but not on the sampled action, the subtraction leaves the expected conditional gradient unchanged and can reduce its variance. vOPD uses the conditional mean teacher--student log ratio
\begin{equation*}
 b_t(c_t)=\mathbb{E}_{a\sim p_\theta(\cdot\mid c_t)}
 \left[\log\frac{p_T(a\mid c_t)}{p_\theta(a\mid c_t)}\right]
 =-D_{\mathrm{KL}}\!\left(p_\theta(\cdot\mid c_t)\parallel p_T(\cdot\mid c_t)\right)
\end{equation*}
as this detached baseline, also called a control variate~\citep{2605.07865}. A top-$K$ approximation retains the baseline property when it remains independent of the sampled action. It approximates the centering value without removing actions from the objective. This baseline centers immediate conditional reward, whereas a value function for cumulative return also accounts for future discrepancies.

Head--tail estimators instead combine exact supervision on a small token set with a sampled correction for the omitted vocabulary. Tail-Corrected Top-$K$ On-Policy Distillation (TT-OPD) sums the reverse-KL gradient over a selected head, using the student's top tokens by default, and adds a correction when the sampled student token falls in the tail~\citep{arxiv260934447}. ResOPD retains the teacher's top tokens and a bucket for the remaining probability mass, then corrects this coarse comparison with a sampled fine-tail residual~\citep{arxiv261004882}. Both recover the full-vocabulary conditional gradient in expectation under their sampling assumptions. ResOPD's control variate reduces variance only when it approximates the tail mean sufficiently well, and its unbiasedness result assumes sampling from the full current student distribution rather than truncated decoding. These constructions distinguish reducing the number of explicit loss terms from discarding tail supervision; neither distinction alone establishes an end-to-end training speedup.

Bounding an update differs from centering it. PowerOPD replaces the detached log ratio with a difference of powered teacher and student probabilities, preserving its sign while bounding its magnitude~\citep{2606.17199}. Because the exponent changes which probability ranges receive substantial updates, it modifies the learning signal rather than providing an unbiased estimate of the original reverse-KL gradient. TOP-D bounds the signal through target construction instead. Mixing the external teacher with the rollout student's probabilities forms a proximal teacher. Its log-probability reward relative to that rollout policy is bounded below, and clipped importance-ratio updates reuse each generated batch~\citep{2607.04751}. This probability mixture differs from scaling a log ratio. In practice, TOP-D holds advantages fixed and normalizes returns, distinguishing its implemented update from the ideal policy-improvement analysis.

Estimator reliability also differs from update magnitude and task usefulness. SuRe emphasizes surprising sampled tokens, motivated by a gradient-norm analysis of its sampled squared-log-ratio surrogate~\citep{2608.25643}. That ranking need not hold for full-vocabulary KL or JSD. IER instead estimates signal relative to single-sample gradient noise in Fisher geometry under an optimized scalar baseline~\citep{2609.24432}. A reliable estimate can still describe a small or unhelpful update; accordingly, IER can be combined with usefulness scores for token selection. The training-cost implications of selecting these tokens depend on which operations are avoided (Section~\ref{subsec:systems}).

Several controls act after a signal has been constructed. PGPO caps a batch-level entropy-gap multiplier without changing optimizer moments~\citep{2606.03620}, while IDA-OPD shrinks sampled updates predicted to contract entropy, especially near teacher--student agreement~\citep{2608.29846}. These mechanisms regulate the constructed signal's scale or predicted entropy effect rather than estimator bias.

Figure~\ref{fig:update_choices} separates vocabulary coverage of supervision, differentiation, and temporal credit. The estimators above primarily concern the update at a visited state. The next section extends this view to actions that affect the states visited later in a sequence.

\input{figures/chapter_maps/02_update_choices}

\subsection{Local Supervision and Temporal Credit}
\label{subsec:temporal_credit}

\textbf{Learning at student states does not by itself assign credit for reaching them.} Local conditional matching treats a visited prefix as fixed. A sequence objective also accounts for how an earlier action changes the distribution of later prefixes. MiniLLM derives this distinction from sequence-level reverse KL. For a fixed teacher with support on student-generated responses, its ideal student-sampled gradient can be written as
\begin{align*}
 \nabla_\theta D_{\mathrm{KL}}(p_\theta\parallel p_T)
 &=-\mathbb{E}_{y\sim p_\theta}\!\left[
 \sum_{t=1}^{|y|}(R_t-1)\nabla_\theta\log p_\theta(y_t\mid c_t)\right],\\
 R_t&=\sum_{t'=t}^{|y|}
 \log\frac{p_T(y_{t'}\mid c_{t'})}{p_\theta(y_{t'}\mid c_{t'})}.
\end{align*}
The constant score term vanishes in expectation, while $R_t$ assigns an early action credit for later teacher--student discrepancies~\citep{2306.08543}. MiniLLM computes the immediate conditional component over the vocabulary and estimates the future component. Its practical update uses mixed teacher--student sampling, per-token importance ratios in place of trajectory weights, and length normalization. Without teacher-mixed sampling or length normalization, reverse KL can still decrease while the student quickly learns repeated, short, or meaningless responses.

Two counterexamples address reward alignment and temporal credit~\citep{2606.05152}. Even when the teacher has higher expected reward, an idealized reverse-KL natural-gradient step can decrease student reward in a finite-action construction. In a restricted policy class, a local token-wise gradient can also assign zero credit to an early decision whose effect appears only in later disagreement, while the full sequence gradient changes it. These existence results rule out unconditional monotonic-improvement and local-credit guarantees.

$\gamma$OPD exposes the temporal horizon directly. With
$\Delta_t=\log p_\theta(y_t\mid c_t)-\log p_T(y_t\mid c_t)$, it defines
\begin{equation*}
 A_t^{(\gamma)}
 =-\sum_{t'=t}^{T}\gamma^{t'-t}\Delta_{t'} .
\end{equation*}
The endpoints recover local credit at $\gamma=0$ and an undiscounted return-to-go at $\gamma=1$~\citep{2609.16937}. Under a uniform second-moment assumption, fixed $\gamma<1$ gives a horizon-independent upper bound on the variance of this advantage estimate, rather than the complete parameter-gradient estimator. In the sensitivity study without reward mixing, undiscounted credit produces much larger gradient norms and rapid entropy collapse, whereas an intermediate discount performs better. Bounded normalization and verifier reward preserve the response-level reward sign while teacher credit changes token-level magnitude. Students can exceed reported teacher scores in the same-size and multi-teacher settings, but the size-mismatched student remains below its teacher.

Other approaches modify the evidence used to assign credit without reproducing this return. X-KD learns a reward distribution and encourages consistency between that distribution and temporal differences of the student's value estimates. Its gains vary by teacher--student setting and auxiliary weight~\citep{2602.12674}. CREDIT instead compares a teacher's feedback-conditioned predictions for the current input with its predictions under other inputs in the batch~\citep{2605.11613}. Subtracting this contrastive baseline reduces credit shared across unrelated inputs. Because the subtraction depends on the candidate action, it changes the supervision signal rather than providing the action-independent centering used by vOPD. This interpretation requires feedback compatible with the teacher's conditioning.

\subsection{From Optimization to Learning Outcomes}
\label{sec:landscape}
\label{subsec:design_axes}

These foundations leave three questions for the rest of the survey. What makes guidance useful at a student's state, given that a stronger teacher or richer context need not transfer more effectively~\citep{2604.13016,2605.00642}? How should a matching objective relate to task rewards when their interaction depends on whether learning is joint or sequential~\citep{2609.04108}? And which outcomes should count as a gain? Response accuracy, coverage of correct solutions, calibration, termination, and readiness for subsequent learning measure different outcomes~\citep{2606.26091,2609.28145}. Pass@$k$, for example, measures the chance of obtaining a correct answer within a budget of $k$ attempts.

The chapter map in Figure~\ref{fig:taxonomy} connects these questions to the different uses of OPD. The next chapter begins with the target: what information a supervisor can provide, when its predictions remain reliable, and how that information can be expressed at the student's interface. These choices determine the supervision available to the objectives above.

%% file: figures/opd_loop.tex
\begin{figure}[!htb]
\centering
\begin{tikzpicture}[
    x=1cm,y=1cm,
    stage/.style={fill=#1,draw=none,rounded corners=3.5pt,
      minimum width=4.18cm,minimum height=3.45cm,inner sep=0pt},
    heading/.style={font=\fontsize{10}{12}\selectfont\bfseries,anchor=west},
    body/.style={font=\fontsize{9}{12}\selectfont,align=left,anchor=north west,
      text width=3.80cm,inner sep=0pt},
    note/.style={font=\fontsize{8.5}{10.5}\selectfont,text=figuremuted,align=center},
    arrow/.style={-{Stealth[length=1.9mm,width=1.35mm,inset=0.4mm,round]},
      line width=0.7pt,line cap=round,line join=round,draw=midgray}
]
\definecolor{figureink}{HTML}{292C2E}
\definecolor{figuremuted}{HTML}{596263}
\definecolor{deeporange}{HTML}{945C43}
\definecolor{deeppurple}{HTML}{6B5C78}
\definecolor{deepblue}{HTML}{3D6871}
\definecolor{midgray}{HTML}{636F72}
\definecolor{looporange}{HTML}{F4ECE6}
\definecolor{looppurple}{HTML}{F0ECF2}
\definecolor{loopblue}{HTML}{EAF1F1}
\tikzset{text=figureink}
\node[stage=looporange] (states) at (2.09,0) {};
\node[stage=looppurple] (teacher) at (6.94,0) {};
\node[stage=loopblue] (update) at (11.79,0) {};
\foreach \x/\c in {0/deeporange,4.85/deeppurple,9.70/deepblue}
  \draw[\c,line width=1.1pt,line cap=round] (\x+0.19,1.725) -- (\x+3.99,1.725);
\node[heading,text=deeporange] at (0.19,1.28) {Collect experience};
\node[heading,text=deeppurple] at (5.04,1.28) {Construct supervision};
\node[heading,text=deepblue] at (9.89,1.28) {Update the student};
\node[body] at (0.28,0.82) {Which states are visited?\\[5pt]
  Coverage and selection\\Refresh, reuse and stages\\Computation budget};
\node[body] at (5.13,0.82) {What defines the target?\\[5pt]
  Adaptation; self-teaching\\Rewards and specialists\\Information access};
\node[body] at (9.98,0.82) {Which gradient is used?\\[5pt]
  Matching; token coverage\\Variance; temporal credit\\Update constraints};
\node[note,anchor=west] at (0.19,-1.38) {Section~\ref{sec:dynamics}};
\node[note,anchor=west] at (5.04,-1.38) {Sections~\ref{sec:signal}--\ref{sec:mopd}};
\node[note,anchor=west] at (9.89,-1.38) {Section~\ref{sec:objectives}};
\draw[arrow] (states.east) -- (teacher.west);
\draw[arrow] (teacher.east) -- (update.west);
\draw[arrow,draw=deepblue,rounded corners=7pt]
  (update.south) -- (11.79,-2.22) -- (2.09,-2.22) -- (states.south);
\node[note,fill=white,inner xsep=5pt,text=deepblue] at (6.94,-2.22)
  {The updated student changes the next state distribution};
\node[note] at (6.94,-2.70)
  {Interaction and memory (Section~\ref{sec:agentic}) $\cdot$
   Behavioral evaluation (Section~\ref{sec:understanding})};
\end{tikzpicture}
\caption{The OPD learning loop. Student-generated states determine where supervision is requested, a teacher or constructed target supplies guidance, and the update changes subsequent student behavior. Self-distillation, task feedback and specialist integration modify the supervision within this loop. Agentic interaction additionally changes observations and visible memory.}
\label{fig:opd_loop}
\end{figure}

%% file: figures/kl_comparison.tex
\begin{figure}[!htbp]
\centering
\includegraphics[width=\textwidth]{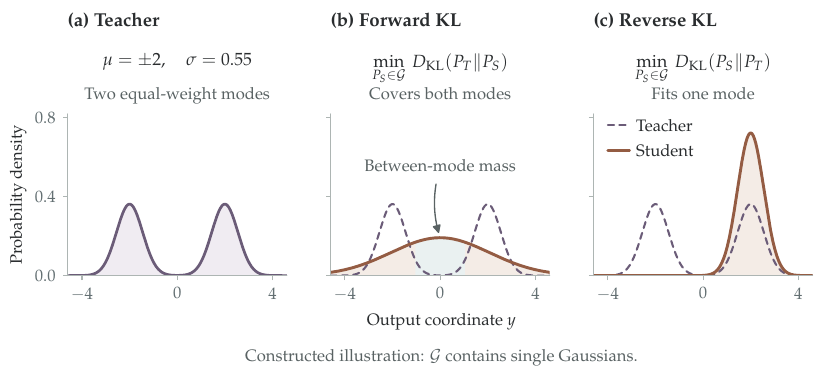}
\caption{Forward and reverse KL in a constructed distribution-fitting example. The teacher is an equal-weight Gaussian mixture, and the student is restricted to a single Gaussian. Forward KL yields the moment-matching fit, covering both modes and assigning more mass between them. Numerical reverse-KL optimization concentrates near one mode for this separation; the reflected solution is equally good.}
\label{fig:kl_divergence}
\end{figure}

%% file: figures/chapter_maps/02_update_choices.tex
\begin{figure}[!htb]
\centering
\includegraphics[width=\linewidth]{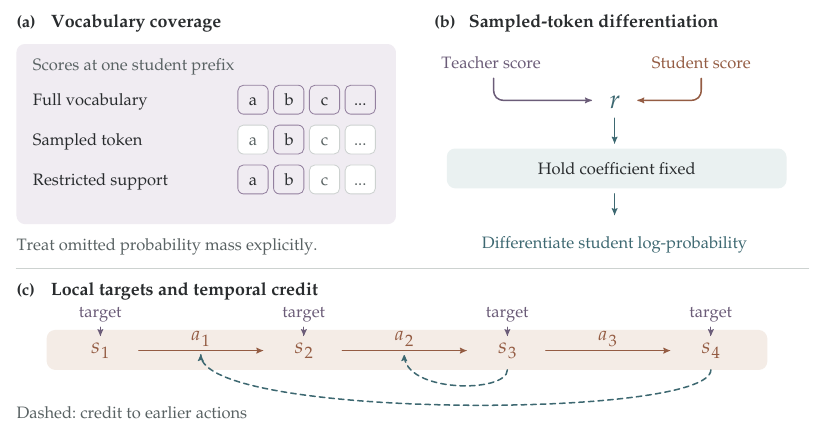}
\caption{OPD update choices. (a) Highlights identify teacher scores included in the loss, rather than student logits receiving gradients. (b) A sampled-token surrogate holds the teacher--student coefficient fixed while differentiating the student log-probability. (c) Local targets supervise visited states; dashed arrows additionally assign credit to earlier actions for their effects on later states.}
\label{fig:update_choices}
\end{figure}

%% file: chapters/03_teacher.tex
\section{Teacher Supervision and Target Construction}\label{sec:signal}

Learning at student states brings the teacher to the relevant history, but useful supervision still depends on what it can provide there. A teacher can become unreliable after a student error, teach behavior outside the intended transfer, or predict tokens the student cannot express. \textbf{A useful target must connect available teacher information to a decision the student can learn.} In Equation~\ref{eq:common_match}, this connection runs from the teacher's view $s^T$ to the target distribution $q$ at the student's state. Supervisory access, teacher reliability, target construction, and prediction-interface alignment determine whether that connection provides effective guidance.

\subsection{Supervisor Roles and Access}\label{subsec:teacher_roles}

A supervisor can supply an answer, judge a response, or predict a distribution at a student prefix. These roles expose different information. A generated answer supplies one example; a judgment evaluates the realized response; a conditional distribution also scores alternatives the student did not choose. The available interface therefore constrains which form of supervision can be constructed.

GKD uses \emph{white-box access}, exposing teacher scores over a shared vocabulary for both chosen and unchosen tokens at student-generated prefixes~\citep{2306.13649}. These scores reveal how probability could be redistributed among alternatives, beyond judging the sampled token alone. The resulting supervision requires teacher inference on the prefixes used for matching, with its cost considered in Section~\ref{subsec:systems}.

Teacher logits also support ranking-based supervision. TSD-KD uses teacher scores to rank the student's candidate next tokens, concentrating ranking supervision on an entropy-determined opening segment alongside distribution matching~\citep{2603.13260}. The ranking term transfers relative preferences among plausible actions, complementing the probability-matching objective.

With text-only access, supervision must recover useful information without the missing conditional distribution. One route transfers generated answers and rationales as training examples~\citep{2305.02301,2501.12948}. Continuing a student-written prefix, as in DASD, makes those examples responsive to student history, although the resulting mixed-policy corpus remains offline during fitting~\citep{2601.09088}. A second route uses teacher demonstrations to evaluate fresh student outputs: GAD and PRISM learn discriminators whose rewards evolve with the student~\citep{2511.10643,2604.28123}. A third makes the criteria explicit. ROPD derives prompt-specific rubrics from teacher--student response contrasts and rewards criterion satisfaction~\citep{2605.07396}. These routes transfer examples, learned judgments, and explicit criteria, respectively. The latter two preserve responsiveness to the current student through a response-level reward, without reconstructing a conditional token target.

Obtaining evidence and converting it into a target can involve different supervisors. SDPO conditions a self-teacher on execution feedback, while On-Policy Critique Distillation uses a frozen, critique-conditioned copy of the student to convert accepted weak-critic comments into token supervision~\citep{2601.20802,2606.00424}. Target construction can also separate the behavior to transfer from the behavior to retain. Direct-OPD uses pre- and post-RL teacher policies to define a relative improvement while the student's initialization anchors behavior~\citep{2607.05394}. The model that supplies evidence, the model that predicts a target, and the policy that anchors the update can therefore play distinct roles.

\subsection{Teacher Reliability and Adaptation}\label{subsec:teacher_validity}

\textbf{Teacher reliability is conditional on the student's history.} OPD deliberately exposes the teacher to prefixes it did not generate, including incorrect reasoning steps. Supervised initialization and prompt format affect transfer~\citep{2604.13016}. In mathematical reasoning, a fixed teacher's continuation accuracy declines as student prefixes lengthen~\citep{2605.30833}. Under reference-conditioned self-distillation, low token-averaged KL can also coexist with incorrect trajectories~\citep{2606.08432}. Comparing raw student rollouts with teacher-refined trajectories further reveals changes in the supervision induced by their histories. Standalone teacher accuracy can therefore conceal weaknesses at the states where distillation occurs.

Improving supervision can require changing the teacher itself. Distillability calibration uses task rewards, an original-teacher anchor, and a smaller proxy model to prepare a teacher for transfer~\citep{2604.18963}. Prefix-conditioned adaptation addresses a more specific weakness: the teacher must recover from histories it would not generate. SCOUT periodically trains on prefixes from the evolving student, whereas Prep-OPD prepares a supervisor on initial-student prefixes and then freezes it~\citep{arxiv260938360,arxiv261004950}. Their comparisons support the value of adapting to student history, including gains from Prep-OPD's prepared teacher without its additional generation handoffs. The choice is whether to track the evolving student's generated prefixes throughout training or obtain a reusable supervisor before distillation begins.

Eligibility testing selects whether the teacher's guidance enters the student update. TGOPD probes teacher rollouts and routes each prompt to distillation or reward learning~\citep{2609.02998}. Confidence does not consistently predict reliability across the tested code and mathematics settings, yet setting the admission threshold too high also degrades performance. Reliable guidance and sufficient training coverage are competing concerns for the gate.

Local target repair changes the prediction after the teacher encounters a student history. In OPD-Aha's study, retaining more erroneous history shifts a crop-conditioned teacher's length-normalized likelihood preference from the correct answer toward the student's realized wrong answer~\citep{2609.16459}. Contrasting predictions under a real privileged image and a null image strengthens visual evidence in the target. Unlike teacher preparation or eligibility gating, this repair preserves the supervisor and training state while changing the distribution used to teach the student.

\subsection{Transformed and Relative Targets}\label{subsec:target_construction}

\textbf{The teacher's distribution and the student's target need not be identical.} Target construction specifies which behavior to transfer, while the fitting divergence specifies how the student approaches it. A relative target makes this distinction concrete. At a shared prefix, start with a reference model's logits, add a scaled difference between two models' logits, and normalize with softmax. The reference supplies the baseline preferences, the contrast changes the relative preference for candidate tokens, and the scale controls the strength of that change. W2S-OPD uses the student's base policy as the reference and a stronger-minus-weaker contrast from two smaller models~\citep{2607.26246}. All three models are frozen while the current student learns from the constructed target, as illustrated in Figure~\ref{fig:teacher_targets}(a).

\input{figures/chapter_maps/03_teacher_targets}

The meaning of the contrast determines which behavior is transferred. Whereas W2S-OPD contrasts stronger and weaker models, Direct-OPD isolates a teacher's post-RL versus pre-RL log-probability change~\citep{2607.05394}. Both anchor the change in the student's base policy, retaining a role for its existing behavior instead of adopting a contrast model's complete distribution. The contrast can also compensate for model-family differences, as in CompassOPD, or compose several improvements to balance accuracy and response length, as in Lightning Weave~\citep{2609.10154,2609.14708}. Choosing the models being contrasted thus determines which part of the teacher's behavior counts as an improvement and which part is subtracted away.

The strength of the requested change is a separate choice. A target can favor agreement between teacher and student or extrapolate beyond a reference. Veto takes the former approach by multiplying teacher probabilities by a powered student distribution and normalizing the product, holding the resulting target fixed during the gradient step~\citep{2601.07155}. G-OPD instead scores trajectories with a scaled teacher-to-reference log-probability ratio and penalizes divergence from the reference~\citep{2602.12125}. At unit scaling, its objective corresponds to reverse-KL matching toward the teacher under the relevant support conditions. Larger scaling extrapolates beyond the teacher along its log-probability difference from the reference, without adding an independent task reward. In the evaluated mathematics and code settings, extrapolation can help, but excessive scaling accompanies degraded performance and longer responses.

Sharpening a complete sequence distribution differs from lowering the temperature of each conditional prediction. OPPD targets a powered teacher sequence distribution, using student proposals inside sequential Monte Carlo and teacher scores to construct both sampling and training weights~\citep{arxiv261006804}. The student learns through detached weighted likelihood together with a token-level forward-KL anchor. The search is performed during training, while deployment uses the resulting student directly. Finite self-normalized importance sampling and the additional anchor make the implemented update an approximation to the ideal powered target, rather than exact sequence-distribution matching.

Evolving policies expose a tension between anchoring to established behavior and extrapolating beyond it. SR-OPSD combines an exponential moving average (EMA) teacher with a frozen reference through a normalized weighted average of log probabilities: one supervisor tracks learning while the other holds a fixed policy. WDL-OPD instead co-trains two policies by matching their geometric mixture to a frozen teacher with reverse KL~\citep{2608.09745,2608.09447}. RISE constructs its teaching policy by extrapolating an RL update direction in weight or logit space~\citep{2609.05295}. The comparisons show why the strength of either intervention must be considered with the learning process. SR-OPSD's anchor helps one divergence setting and hurts another; RISE collapses without reward-learning updates, and aggressive extrapolation becomes harmful later in training. A target that is useful at one stage need not remain so as the policies evolve.

The preceding contrasts compare policies, but different evidence supplied to the same teacher can also define a useful change. OGLS-SD contrasts successful and failed guidance pools to suppress artifacts shared by privileged prompts~\citep{2605.12400}. Visual Gradient Steering (VGS) contrasts the teacher's image-conditioned and text-only predictions at multimodal student prefixes~\citep{2606.00564}. It combines this log-probability difference with the student's text-only prior to form a locally normalized target, then adds visual-target matching and selective language preservation to ordinary OPD. Steering toward the visual target improves the tested visual-reasoning results, whereas steering toward language matching lowers accuracy. The contrast operates on conditional predictions, without an exact decomposition of sequence-level knowledge.

A likelihood contrast can also enter as a reward without defining a single matching target. COPD scores the same student token under light- and heavy-thinking teacher contexts and uses the clipped, detached light-minus-heavy log-likelihood difference as an advantage~\citep{arxiv260719046}. Before clipping, its fixed-state objective is a difference of two KL divergences: the student entropy terms cancel. This differs from fitting a normalized relative target. The reported recipe favors shorter responses with a task-dependent accuracy tradeoff, connecting contrastive guidance to the behavior being transferred rather than to teacher imitation alone.

Teacher context can support more direct evidence about a continuation or an instruction. AC-OPD generates supervised continuations from selected student prefixes, while CC-OPD compares scores under the full instruction and contexts that omit one constraint~\citep{2609.22254,2609.27421}. The former adds teacher-generated targets alongside the OPD loss on the full student rollout; the latter asks which instruction makes a decision preferable. Both incur additional teacher work. Longer lookahead is not uniformly beneficial, while constraint-wise scoring grows with the number of instruction views. The value of additional evidence therefore depends on how it changes the guidance the student receives, as well as on the cost of obtaining it.

\subsection{Aligning Prediction Interfaces}\label{subsec:interfaces}

\textbf{Teacher and student probabilities must refer to comparable events.} The shared prefix and vocabulary in the relative-target example make its logit contrast well-defined. With different tokenizations, the same text can instead span different prediction steps, as Figure~\ref{fig:teacher_targets}(b) illustrates. Interface methods address this mismatch at the level of continuations, byte prefixes, sequences, or representations, trading conditional probability information for compatibility with broader model pairs.

The first difficulty is deciding when differently tokenized predictions refer to the same event. Shared short continuations and aligned text chunks broaden the positions that can be supervised, but matching a sampled chunk's likelihood does not establish normalized matching over alternative continuations~\citep{2605.07711,2606.09456}. BPM instead groups teacher probability by student byte prefixes and explicitly accounts for unmatched mass and stopping~\citep{2607.22334}. Its exactness depends on prefix and refinement assumptions; some spanning-token corrections remain lower bounds and unsupported cases are excluded. Even a more faithful mapping can require behavioral safeguards: its byte-exact whitespace targets cause an indentation collapse in one code setting unless those positions are excluded.

Broader coverage is therefore useful only when the added correspondence is reliable. A cross-tokenizer analysis finds that adding span-score matching to strict shared-token matching degrades performance across its three tested model pairs~\citep{arxiv261008448}. Text-only access relaxes correspondence further. OmniOPD compares teacher continuations with uncertain student chunks through an expected-similarity proxy~\citep{2606.01476}. This permits cross-tokenizer guidance without teacher logits, but the proxy is not a teacher chunk probability, and its comparison with white-box OPD changes with the similarity metric and model pair. These results expose a common tradeoff: more positions or broader teacher access can come at the expense of a less precise supervision target.

Sequence-level objectives avoid token alignment by changing the supervised object. GVPO++ matches group-centered teacher--student differences in sequence log probability, with both matched- and cross-tokenizer teachers~\citep{2609.21432}. In contrast, relative-logit methods such as CSD retain a shared-vocabulary requirement~\citep{2509.25837}. The broader tokenizer compatibility comes from matching sequence scores instead of token-conditional distributions.

Representation matching trades token correspondence for a different compatibility problem. OPRD bridges hidden states across architectures, with more favorable results within the same architecture~\citep{2606.06021}. LastOPD shows why successful latent alignment alone is insufficient: in its cross-size pairs, task accuracy eventually collapses even as projected cosine similarity improves~\citep{2609.28845}. Transitioning to token supervision helps those pairs, while continued latent supervision works better for a same-lineage pair. The relevant question is thus whether a closer representation continues to support useful output behavior, and for how long.

Cross-modal supervision must additionally account for what each model can observe. CORD compares audio- and text-conditioned predictions on audio-policy prefixes through weighted token alignment and a sequence-level consistency judge~\citep{2601.16547}. FP-OPD asks whether a visual correction lies within a student's locally probed response space, Fisher-projecting the correction before fitting it~\citep{2608.01263}. The probes add forward passes and characterize a local response space, not global parameter reachability. Both methods connect output compatibility to the student's available evidence.

Diffusion generation makes the conditioning state part of the alignment problem. Hindsight predictions at later masked states can supervise earlier decisions; alternatively, intermediate decisions retained in a final response can be selected for teacher supervision at their original denoising states~\citep{2607.04428,2607.16872}. Across models with different block sizes, however, the same output position can be predicted from different committed tokens. Un-OPD filters incompatible commitment states and calibrates high-support targets toward the student to curb its premature overconfidence~\citep{arxiv261005373}. Matching positions is therefore insufficient: the visible contexts must also be comparable. Filtering and calibration change the supervised objective rather than establishing an unbiased estimate of the original one.

Distributional, semantic, and latent targets preserve different information about the teacher's behavior. A teaching advantage can also arise within a single model when its views differ. The next chapter examines how demonstrations, feedback, and informative views let a model construct supervision from its own parameters.

%% file: figures/chapter_maps/03_teacher_targets.tex
\begin{figure}[!htb]
\centering
\includegraphics[width=\linewidth]{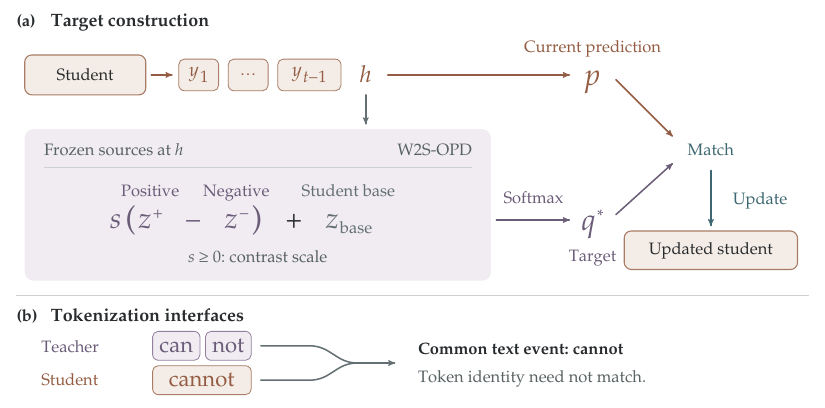}
\caption{Teacher signals and the target to be fitted. (a) W2S-OPD adds a scaled positive-minus-negative logit contrast to the student base logits, using three frozen models to construct the target at a shared prefix~\citep{2607.26246}. Only the current student is updated; both contrast models can be weaker than the student base. (b) The same text can have different token boundaries. The schematic correspondence illustrates why segmentation alignment alone does not establish a probability-preserving map~\citep{2605.07711,2606.09456,2607.22334}.}
\label{fig:teacher_targets}
\end{figure}

%% file: chapters/04_self_distillation.tex
\section{On-Policy Self-Distillation}\label{sec:self_distillation}

A model can sometimes judge a continuation better than it can generate one unaided. A worked solution, demonstration, test failure, or privileged observation changes the information available for that judgment; additional internal computation can provide another teaching advantage. \textbf{On-policy self-distillation turns these advantages into supervision at the student's own states.} The supervisor comes from the student model or its parameter snapshots, which need not remain identical to the evolving student. In context-conditioned methods, it evaluates the student's attempts with extra information, while the student learns to act from the deployment input alone. This offers an alternative when a suitable stronger teacher is unavailable or costly, although constructing and evaluating a useful self-teacher still requires resources.

For context-conditioned self-distillation, a common local objective is
\begin{equation}
\loss_{\mathrm{self}}(\theta)=
\mathbb{E}_{x\sim\mathcal D,\,y\sim\pi_\theta(\cdot\mid x),\,c\sim\mathcal C(\cdot\mid x,y)}
\left[\sum_t D_f\!\left(
\operatorname{sg}[q_{\bar\theta}(\cdot\mid x,c,y_{<t})]
\parallel\pi_\theta(\cdot\mid x,y_{<t})\right)\right].
\label{eq:self_match}
\end{equation}
Here $\mathcal C$ supplies teacher-only context $c$, either before generation, as with a demonstration, or after observing the attempt $y$, as with feedback. The teacher re-scores the student's prefixes, with $\operatorname{sg}$ stopping gradients through the target. Differentiation through the sampled-state distribution remains a separate choice (Section~\ref{subsec:estimators}). The teacher parameters $\bar\theta$ may be frozen, periodically synchronized, or updated by an exponential moving average (EMA). Teacher-only information ranges from solutions, demonstrations, and retrospective feedback~\citep{2601.18734,2601.19897,2601.20802} to reusable instructions and learned criteria~\citep{2602.12275,2606.19327}. Table~\ref{tab:self_teachers} compares how these sources of information are combined with teacher evolution and matching signals.

Figure~\ref{fig:self_teacher} makes the information asymmetry explicit. Additional context changes the self-teacher's evaluation of the student's existing attempt, while the updated student retains the deployment input interface.

\input{figures/chapter_maps/04_self_teacher}

\begin{table}[t]
\centering
\small
\setlength{\tabcolsep}{4pt}
\renewcommand{\arraystretch}{1.08}
\begin{tabularx}{\linewidth}{@{}P{0.10\linewidth}P{0.23\linewidth}P{0.19\linewidth}>{\raggedright\arraybackslash}X@{}}
\toprule
\textbf{Method} & \textbf{Teacher-only information} & \textbf{Teacher evolution} & \textbf{Student states and matching signal} \\
\midrule
\textbf{OPSD} & Reference solution & Frozen initial model & Response prefixes; full-vocabulary forward KL with pointwise clipping. \\
\addlinespace[3pt]
\textbf{SDFT} & Expert demonstration & EMA of student & Response prefixes; full-vocabulary matching. Forward KL favored empirically; reverse-KL formulation. \\
\addlinespace[3pt]
\textbf{SDPO} & Feedback on an attempt & Regularized self-teacher: EMA or initial-teacher interpolation & Original attempt prefixes; feedback-conditioned distributions, with KL and JSD variants. \\
\addlinespace[3pt]
\textbf{OPCD} & Experience or system instructions & Frozen in the main self-distillation experiments; other schedules studied & Response prefixes without privileged context; student top-$K$ approximation to reverse KL. \\
\addlinespace[3pt]
\textbf{RCSD} & Reference answer, then a generated rubric & Fixed within training & Two stages: rubric prefixes, then answer prefixes; forward KL in each. \\
\bottomrule
\end{tabularx}
\caption{Representative context-conditioned self-teachers, distinguished by their privileged information, parameter evolution, and matching signal.}
\label{tab:self_teachers}
\end{table}

\subsection{Solutions, Demonstrations and Reusable Experience}\label{subsec:context_construction}

Context-conditioned methods differ in where privileged information enters learning: it can change the sampled trace, the target at an existing prefix, or the reusable guidance supplied across problems. These choices determine what the student is asked to internalize.

\subsubsection{Where the reference enters supervision}

Worked solutions and demonstrations can inform a self-teacher's judgments at the student's own prefixes, without becoming responses the student must reproduce~\citep{2601.18734,2601.19897}. The benefit depends on whether the model can turn that context into useful guidance. SDFT improves acquisition and retention relative to SFT in its science-adaptation comparison, whereas its smallest evaluated model trails SFT. Retention is also incomplete in medical adaptation, where some prior instruction-following performance degrades. These contrasts connect the value of a teaching view to in-context competence and the capabilities being preserved. SDFT reports better performance with forward KL in practice, while its analysis uses reverse KL.

Privileged information can enter elsewhere in learning, with different consequences for state coverage. DAIL and Hinted Decoding use it to construct fixed training traces, while SAPD uses contextual soft targets at fixed reference prefixes~\citep{2602.02405,2602.12222,arxiv261009665}. These designs improve the examples or targets available for teacher-forced learning without continuously refreshing student trajectories.

Student-generated responses create further uses for the reference. GATES combines teacher-trajectory imitation with scoring of student rollouts, gating both through repeated-answer consensus~\citep{2602.20574}; that consensus supplies a reliability proxy rather than verified correctness. PBSD changes the supervised object instead, preferring independently generated privileged responses over current-student responses~\citep{2605.05040}. Thus, access to the same kind of reference can support trace construction, conditional correction, or response-pair preference learning. These routes differ in whether student prefixes are refreshed and how informed predictions enter the update.

\subsubsection{Information access and target construction}

Once the teacher evaluates an existing prefix, the next question is which information should inform that evaluation. A comparison separates teacher-sampled $\pi$-Distill from student-sampled OPSD and finds that privileged traces have different value across models, tasks, and objectives~\citep{2602.04942}. PAINT favors moderate suffix masking in its disclosure comparison, while Adaptive Teacher Exposure learns the amount exposed~\citep{2604.26573,2605.11458}. \textbf{More privileged information is not reliably more transferable information.} A complete path can be informative yet poorly matched to a student that has taken another route. Partial disclosure therefore changes the teacher's dependence on that path as well as the prompt length.

The teacher's advantage can also come from easier access to the relevant information. OPSDL gives an evolving self-teacher a short evidence context while the student answers from a long document~\citep{2604.17535}. Its long-context gains coexist with small declines on the reported short-context knowledge and reasoning benchmarks. COPSD crosses a language boundary instead, using an English problem and solution to inform a frozen self-teacher that evaluates target-language reasoning~\citep{2605.09548}. It improves low-resource math averages for separately trained language-specific students, rather than a jointly trained multilingual model.

Different contextual views need not express the same desired behavior. SELFCI combines the reverse-KL objectives of two teachers conditioned on allowed and disallowed attribute feedback~\citep{2605.20258}. Its privacy--utility tradeoff depends on those supplied decisions and carries no formal privacy guarantee. OPCT distinguishes variation to ignore from information to retain, using clean/perturbed prompt pairs for invariance and teacher-only safety facts for internalization~\citep{2605.21834}. These constructions give the views different roles before their guidance reaches the student.

Combining views requires deciding what their agreement means. Geometric consensus can suppress a useful token if even one view assigns it little probability. AVSD supplements consensus with a gated arithmetic--geometric residual, allowing support from individual views to influence the update~\citep{2605.20643}. AR-OPD instead assigns the views an explicit direction: it scales the log-probability shift from a partial-information anchor toward full-information predictions~\citep{2606.10385}. The former recovers guidance lost by agreement-based aggregation; the latter uses information asymmetry to define the desired change.

The student must ultimately act without those privileged views. DAPD addresses this gap by aligning predictions with and without completion-level information along both reference and student trajectories~\citep{2608.01735}. Its additional prediction branches connect privileged supervision to the deployment view, at extra computation cost. This differs from aggregating several informed teachers, as in AVSD, and from teaching invariance to input perturbations, as in OPCT: the central issue is whether guidance constructed with extra information remains learnable when that information is absent.

\subsubsection{From worked solutions to reusable guidance}

Changing access to a worked solution still leaves its content tied to one problem. When guidance comes from accumulated experience, extracting what transfers becomes part of the learning problem. OPCD transfers experience and system instructions using either external teachers or teachers initialized from the same model~\citep{2602.12275}. It improves test accuracy over off-policy context distillation in most evaluated configurations, and its experience-consolidation comparisons retain more out-of-distribution instruction-following performance. Yet prepending raw traces degrades math validation performance, whereas extracted lessons help. This comparison motivates separating the experience itself from the guidance derived from it.

Guidance can specify a procedure or a criterion. Harness-generated drafting and verification context, problem-solving structures, and skill cards describe how to proceed~\citep{2605.08741,2608.01589,2608.09826}; rubrics describe what a satisfactory answer should achieve~\citep{2606.12507,2606.19327,2608.02948}. They can replace a full worked solution in the reasoning teacher's immediate context, even when reference solutions are used to train the guidance generator. Unlike response-level rubric rewards such as ROPD (Section~\ref{subsec:teacher_roles}), rubric-conditioned self-teachers convert criteria into token predictions. RCSD further learns to generate the rubric itself~\citep{2606.19327}. Yet generic and randomly reassigned rubrics also improve over its base model, cautioning against attributing every gain to precise instance-specific criteria.

A compact lesson is useful only if its value transfers beyond the attempt that produced it. SPEE tests lessons on held-out probe problems before transferring them, whereas SGUID selects skills whose distillation signals remain aligned with verified outcomes as training progresses~\citep{2608.02139,arxiv261012367}. These approaches test different forms of persistence: usefulness across problems and usefulness across student stages. SGUID's selection requires an initial full-bank run before retraining, so a smaller retained bank does not by itself establish a lower total training cost. The broader design problem is to preserve actionable experience without continually carrying the full history that generated it.

\subsection{Feedback and Hindsight}\label{subsec:self_feedback}

\textbf{Feedback can change the evaluation of an attempt without replacing the attempt.} SDPO supplies a runtime error, failed test, or successful attempt when re-scoring the original student prefixes~\citep{2601.20802}. The same token can consequently receive different guidance after its consequences become known. When only scalar outcomes exist, successful attempts can supply contextual feedback.

Revision creates two distinct learning opportunities: imitate the corrected response or reconsider the original attempt with better information. RLTF-SD follows the first route by transferring critique-guided revisions alongside multi-turn RL~\citep{2602.02482}. The second uses revision-informed teachers to re-score the original generator prefixes, with the attempt and its outcome informing the supervisory distribution~\citep{2604.12002,arxiv261011529}. A successful revision can therefore improve supervision without becoming the response directly imitated. Unsuccessful revisions may also explain a mistake, but their value as guidance is task dependent, as ReTeach's comparisons show.

Corrective evidence must then be assigned to the attempts and decisions it can inform. SRPO admits failed attempts with usable corrective information; ROSD localizes supervision to a diagnosed first-error boundary, falling back to the full response when localization fails~\citep{2604.02288,2605.28014}. RESD and Multi-Rollout OPD address a different bottleneck by supplying persistent diagnoses or successful and failed peers on the same prompt~\citep{2605.12741,2605.12652}. Admission and localization determine where to learn; memory and peers determine what evidence is available there. Section~\ref{subsec:selective_training} connects this allocation to reward updates.

Feedback can even determine whether guidance should be followed or reversed. Direction-Adaptive Self-Distillation uses local student entropy to choose attraction or repulsion, with a gap-based gate and verifier reward~\citep{2605.22263}. SGSD instead compares skill-conditioned teacher support with trajectory success, reversing guidance that supports failure or opposes success and ignoring weak support~\citep{2605.28791}. Both relax unconditional imitation, but their decision criteria differ: uncertainty describes the student's local prediction, whereas the outcome evaluates what the trajectory achieved.

Useful feedback requires both an informative message and a model able to interpret it. Improving the message alone can help: extracted coaching advice outperforms raw critiques and rubric-only inputs in LLM-as-a-Coach's format comparisons~\citep{2607.18110}. Other approaches change the model that acts on the evidence. PAST adapts the supervisor to verified student attempts before re-scoring their original prefixes, while SELF couples hindsight distillation with learning to predict environment feedback~\citep{2608.08726,arxiv261011384}. The former balances verifier-guided correction with student compatibility; the latter trains the shared model to anticipate consequences. Their additional objectives, and PAST's teacher sampling, make this a change in how feedback is learned from as well as how it is expressed.

Additional feedback can improve one measure while worsening another. In Multi-Rollout OPD's LiveCodeBench comparison, adding one failed peer to one successful peer raises mean@8 but lowers pass@8~\citep{2605.12652}. Average success across eight attempts improves while the fraction of problems solved at least once declines. Feedback quality also depends on who provides it. A separate study finds that an incognizant tutor is more likely to break an initially correct attempt through its feedback than a cognizant tutor, with the distinction based on the tutor's relative validation-performance gap across domains~\citep{2606.14368}.

Hindsight need not be verbal. dOPSD matches token predictions at a masked diffusion state to a stop-gradient average of predictions from later states of the same trajectory, using only states where the supervised position remains masked~\citep{2607.04428}. Training retains only rollouts with correct final answers, confining this form of hindsight learning to successful denoising trajectories.

\subsection{Privileged Perceptual Views}
\label{subsec:privileged_views}

\textbf{A better teaching view need not produce the most accurate standalone teacher.} GUI-SD finds that a boxed visual hint transfers better than textual coordinates despite lower teacher accuracy in that comparison~\citep{2605.00642}. Vision-OPD gives the teacher an evidence-centered crop while the student receives the full image with spatial guidance, and Thinking Without Images conditions a teacher on annotated regions and an answer while supervising textual inspection traces~\citep{2605.18740,2606.08719}. These views make task evidence more accessible to the teacher at the student's current state.

The teaching view may also reveal information unavailable at decision time. World-Model PI gives an evaluator future video and the final answer to supervise whether the student should invoke or trust a prediction. MGSD gives its teacher symbolic planning state and a reference plan while the student acts from an image~\citep{2606.03603,2606.06076}. MGSD's ablations support both the privileged state and a preparation stage. These teachers use future observations or structured planning information to supervise decisions the student must make without them.

Geometric guidance exposes the tension between richer evidence and dependence on hidden information. SpatialOPSD supplies verified traces with executed geometric observations, while GPD routes textual 3D cues and an answer to a teacher supervising an RGB-based student~\citep{arxiv261011366,arxiv261012355}. GPD's comparisons favor selecting cues over supplying all of them. SpatialOPSD shows why the student must also learn what not to copy: masking trace-derived phrases gives its best accuracy, whereas unlikelihood training more strongly suppresses references to the hidden trace. A useful privileged view should improve decisions grounded in the student's input; reducing surface references alone does not establish that all dependence on unavailable information has disappeared.

Clean teaching views can supervise student responses generated from corrupted images or from external-data contexts containing prompt injections~\citep{2607.23125,2608.21500}. The advantage comes from removing the perturbation; in the image setting, this supplies a distillation signal without ground-truth answers. The shared construction serves different goals: the image-based study reports gains on standard clean-image benchmarks, whereas the security study evaluates resistance to prompt injection.

Making a view informative can require preparing the teacher as well as selecting its input. Limited calibration in a constitutional safety study improves selected safety and helpfulness measures, whereas longer calibration can worsen downstream safety~\citep{2606.03089}. View changes can also determine which supervision to retain. OPD-V contrasts evidence-present and masked views to gate tokens~\citep{2608.05131}. A related use of contextual interventions appears in Cal-OPD, which probes how they change a stronger external teacher's likelihood and uses those changes to estimate a range of teacher-side variation~\citep{2609.21619}. It sets the distillation signal to zero inside this range and retains only the student--teacher discrepancy beyond its boundary. These choices tie the value of a view to the guidance it induces at student states, beyond the quantity of information it contains.

\subsection{Self-teacher Evolution and Internalization}\label{subsec:self_evolution}

\textbf{Teacher evolution and privileged information are separate design choices.} A useful context can be supplied to a frozen model or to a supervisor that changes with learning. Comparisons in demonstration- and feedback-conditioned self-distillation favor regularized evolving teachers over frozen or directly updated alternatives~\citep{2601.19897,2601.20802}. SDFT illustrates the tradeoff: freezing gives a stable but weaker teacher, whereas directly using the changing student is unstable; EMA balances stability with adaptation. OPCD instead finds its frozen teacher better than a continuously updated self-teacher in Sokoban and medical comparisons~\citep{2602.12275}. Because the contextual tasks and teacher constructions differ, these results identify a tension between a stable instructional reference and access to newly acquired competence, rather than one universally best schedule.

An evolving target does not necessarily imply evolving teacher weights. SD-Zero keeps its prepared reviser fixed in the main setting and tests an extension that synchronizes it with the updated student, whereas OPHSD changes the harness context supplied to a frozen initial scorer~\citep{2604.12002,2605.08741}. Reviser synchronization changes the supervisor's parameters, while harness adaptation changes the information those parameters act on.

Sharing parameters couples teacher improvement to student learning more tightly, whether the teaching advantage comes from privileged information or additional computation. JOLT trains a privileged branch with reward and compatibility objectives while distilling its predictions into the student branch; LoopOPD uses terminal-loop predictions to supervise intermediate-loop behavior~\citep{arxiv261010447,arxiv261010623}. JOLT and the dynamic D-LoopOPD variant link supervisor evolution to student updates, whereas base LoopOPD freezes its terminal-loop teacher. This distinction matters for the available guarantees: JOLT's local fixed-teacher reward-consistency argument does not cover the complete shared-parameter optimization, and LoopOPD's task-level improvement bound requires aligned distillation and task gradients with a sufficiently small update.

The coupling between branches also makes the teacher schedule depend on the other losses being optimized. UniSD finds task-dependent contributions from EMA targets, contrastive guidance, and representation matching~\citep{2605.06597}. An evolving teacher can therefore be assessed only in relation to the objectives that change it and the student receiving its guidance.

Even a well-chosen self-teacher can only supervise the states training exposes. A student that omits a relevant fact may never visit the prefixes where a reference develops it. Reference-side supervision complements student-path matching by teaching content under reference prefixes, ranging from full-response anchoring to selected factual corrections~\citep{2606.00147,2607.24771,2608.07935}. The central choices are how broadly to restore reference coverage and how strongly to weight it relative to learning on the student's path.

Reference content must also be recoverable from the deployment input. In personalized generation, MIRROR supplies a reference only to the self-teacher, while MIRROR-F additionally teaches selected reference terms that occur in the student's available context~\citep{arxiv261009795}. This makes a useful distinction between exposing the teacher to an answer and grounding the student's learning in accessible evidence. The task-dependent gains leave both coverage and grounding as constraints on successful internalization.

Internalization can change computation as well as answers. Conciseness-conditioned distillation shortens reasoning, but aggressive compression hurts harder tasks~\citep{2603.05433}. Masked Distillation likewise finds that withholding a complete reference trace from the student helps GSM8K yet sharply hurts graph coloring, where partial masking works better~\citep{2607.22629}. \emph{Shorter reasoning is not evidence of error repair.} Section~\ref{subsec:mechanisms} examines this distinction through studies that separate correct from incorrect responses or isolate self-distillation from other objectives~\citep{2605.06188,2608.04794}.

Because contextual guidance can disagree with measured task success, the next chapter examines how outcome feedback complements or modifies teacher supervision. Section~\ref{sec:understanding} returns to internalization through changes in student reasoning and solution coverage.

%% file: figures/chapter_maps/04_self_teacher.tex
\begin{figure}[!htb]
\centering
\includegraphics[width=\linewidth]{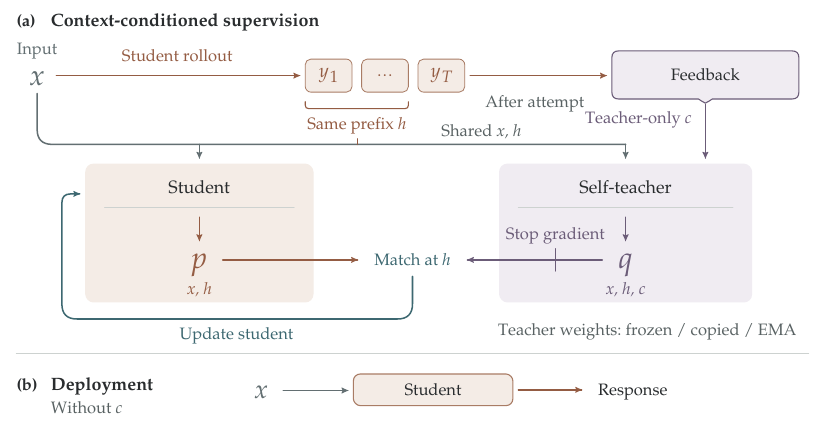}
\caption{Context-conditioned self-distillation. (a) Feedback from the student's attempt supplies extra context $c$ to a self-teacher initialized from the student. Predictions are matched at the original student prefix $h$, with the target held fixed during the update~\citep{2601.20802}. Context construction is separate from the teacher's parameter schedule, which may use frozen, periodically copied, or moving-average weights~\citep{2601.18734,2601.19897}. (b) The deployed student receives $x$ without $c$.}
\label{fig:self_teacher}
\end{figure}

%% file: chapters/05_rl_opd.tex
\section{Combining OPD with Reinforcement Learning}\label{sec:rl_opd}

\textbf{Teacher preference and task success are different learning signals.} A teacher distribution distinguishes plausible next tokens throughout a generated solution, while an outcome reward judges the attempt against a verifier or other task criterion. Their combination is useful when the teacher supplies behavior that helps the student obtain reward. It becomes problematic when teacher-supported tokens belong to an unsuccessful response, or when continued imitation constrains a student that can improve through reward optimization. The design problem is therefore how to combine the signals, decide where each is informative, and change their roles during training.

We distinguish \emph{joint objectives}, \emph{selective supervision}, and \emph{sequential or alternating learning}. These choices concern the update, the allocation of guidance, and the training schedule, respectively, and can coexist in one method. A policy-gradient implementation of distillation, such as MiniLLM or SDPO, does not by itself add an independent task-reward objective~\citep{2306.08543,2601.20802}. Similarly, a reference-policy drift penalty constrains policy movement, whereas OPD supplies a teacher-derived behavioral target on student trajectories.

Figure~\ref{fig:rl_integration} uses one failed rollout to separate the teacher's local preference from task success, then shows how combination, selection, and scheduling act on these signals.

\input{figures/chapter_maps/05_rl_integration}

\subsection{Joint Objectives}\label{subsec:reward_coupling}\label{subsec:joint_training}

GKD combines these two information sources for RL fine-tuning~\citep{2306.13649}. Student-generated outputs supply both the trajectories on which reward is evaluated and the prefixes on which teacher distributions are queried. For a GKD-style update at iteration $k$, let $\mu_k=\pi_{\theta_k}$ denote the rollout policy, held fixed in the matching term. The update objective can be written schematically as
\begin{equation}
\begin{aligned}
\max_{\theta}\quad &\mathbb{E}_{x\sim\mathcal D,\,y\sim\pi_{\theta}(\cdot\mid x)}[r(x,y)]\\
&-\lambda\mathbb{E}_{x\sim\mathcal D,\,y\sim\mu_k(\cdot\mid x)}
\!\left[\sum_t D\!\left(\pi_T(\cdot\mid x,y_{<t}),\pi_{\theta}(\cdot\mid x,y_{<t})\right)\right].
\end{aligned}
\end{equation}
Here $r$ evaluates the attempt and $D$ denotes a chosen conditional matching measure. The reward term uses a policy gradient, while matching treats collected prefixes as fixed. Differentiating through the matching term's state distribution would add the temporal-credit component discussed in Section~\ref{subsec:temporal_credit}. GKD also permits fixed examples to be mixed in, with only the student-generated portion on-policy. Hybrid objectives can thus differ in their divergence, prefix weighting, and treatment of the sampling distribution.

Task evidence can determine either what counts as a discrepancy or how much a separate objective rewards success. The white-box moment-matching branch of PbKD uses preference-constrained value functions to compare teacher actions with student actions at visited states~\citep{2505.18952}. It therefore selects discrepancies through task-relevant preferences instead of imitating the complete distribution indiscriminately. KDRL takes the second route, adding a sampled squared teacher--student log-ratio loss to rule-based reward optimization~\citep{2506.02208}. At a fixed prefix, sampling from the current untruncated student makes the expected gradient of this loss equal the conditional reverse-KL gradient. The squared loss is not itself a KL value, and the identity does not supply future-state credit or automatically survive uncorrected rollout reuse. The two constructions illustrate why the source of task information and the form of the distillation update are separate choices.

The distinction between local matching and trajectory credit remains even when both terms arise from one formal objective. One formulation decomposes trajectory-level student-to-reference KL plus task reward into an analytic local-KL gradient and a sampled gradient carrying future KL costs and terminal reward~\citep{2512.23097}. Unlike a local matching loss added to outcome-based RL, its sampled component retains future KL costs. The analysis locates that temporal-credit contribution, but supplies neither empirical training validation nor a general total-variance guarantee.

Hybrid designs also modify how local teacher evidence interacts with outcome credit. Teacher guidance can accompany long-context reward optimization~\citep{2605.12227}, while CAST uses an answer-free self-teacher to reverse local credit for a teacher-supported token in a failed response~\citep{2606.00172}. CAST changes the \emph{update direction} according to task success. By comparison, relaxed OPD clips a teacher log-ratio within the matching signal, without adding an independent task-reward objective~\citep{2603.11137}.

The teacher can also evolve during joint learning. CoDistill-GRPO couples a changing large-model supervisor with small-model rollouts~\citep{2605.08873}. Section~\ref{subsec:mopd_evolution} distinguishes the target and trajectory exchanges in this asymmetric procedure. Whether the supervisor is fixed or evolving, the next question is which attempts should receive its guidance.

\subsection{Selecting and Weighting Supervision}\label{subsec:selective_training}

\textbf{Selecting failed responses is useful only when corrective information is available.} Group Relative Policy Optimization (GRPO) derives advantages by comparing rewards among responses to the same prompt. Equal rewards provide no within-group preference. A successful sibling response can supply a different kind of evidence. SRPO uses such feedback to correct eligible failed attempts, while SSOPD uses the shortest correct sibling to supervise an initial span of the longest failed attempt~\citep{2604.02288,2605.17497}. The successful response informs a teacher at the failed student's prefixes rather than becoming the directly imitated trace. RLTF-SD instead learns from revised responses, with their contribution determined by reward relative to the first-turn group~\citep{2602.02482}.

This distinction matters most when every attempt fails. Without another feedback source, sibling-conditioned correction lacks a successful example; SSOPD explicitly removes its contribution in all-failure and all-correct groups. An independently supplied teacher target, such as KDRL's, can remain available even when reward contrast disappears. Distill Where You Fail further finds that teacher-based advantage weighting helps before token selection and teacher-response imitation are added~\citep{2608.00782}. Allocation therefore depends on \emph{which learning signal survives}, not simply on whether a response is wrong: group reward contrast, a verified peer, and an external supervisor fail under different conditions.

One response to missing corrective evidence is to search for it before requesting a teacher. CataOPD first allows additional unaided student attempts, reserving teacher guidance for unresolved questions~\citep{2609.29518}. Verified student-produced trajectories then train the unaided policy. Here the teacher helps obtain a successful target instead of supplying a complete policy for matching. The strategy trades additional sampling and verification for more selective teacher use; its correctness-based routing also differs from reward-group comparisons that can vary with formatting even when all answers are wrong.

\textbf{Teacher information can reweight reward credit without becoming a matching objective.} Teacher likelihoods can modulate token-level reward credit while preserving the sign of the reward advantage~\citep{2606.11709,arxiv260717247}. Table~\ref{tab:reward_roles} separates this role from gating teacher updates and constructing residual credit. The distinction is whether the teacher changes the strength of an existing update, selects which updates survive, or helps define a new signal.

\begin{table}[!htbp]
\centering
\small
\setlength{\tabcolsep}{4pt}
\renewcommand{\arraystretch}{1.12}
\begin{tabularx}{\linewidth}{@{}P{0.25\linewidth}P{0.34\linewidth}>{\raggedright\arraybackslash}X@{}}
\toprule
\textbf{Update mechanism} & \textbf{Role of teacher information} & \textbf{Role of task reward} \\
\midrule
Token reweighting\newline RLCSD; Distilled RL; SC-GRPO (reward branch) & Changes the magnitude of reward-derived token credit & Supplies the advantage whose direction is retained \\
\addlinespace
Trajectory gating\newline RG-OPD & Fits the same reverse-KL objective on retained responses & Selects responses jointly with the teacher--student sequence-likelihood gap \\
\addlinespace
Token gating\newline OPDVR & Supplies local update coefficients & Retains coefficients whose direction agrees with correctness or group-advantage sign \\
\addlinespace
Group residual\newline GC-OPD & Supplies a normalized response score and bounded token weights & Forms residual credit by subtracting the distillation score from normalized reward \\
\addlinespace
Distribution residual\newline RA & Supplies a teacher-minus-student probability residual & Combines outcome credit with response-centered teacher guidance \\
\bottomrule
\end{tabularx}
\caption{Different roles for teacher information and task reward in selective supervision. The rows distinguish update mechanisms rather than complete algorithms. SC-GRPO also constructs a diversity pseudo-advantage for all-failure groups; method-specific conditions are discussed in the text.}
\label{tab:reward_roles}
\end{table}

Credit shaping remains constrained by the reward signal it modifies. RLCSD uses a bounded contrast between correct- and incorrect-hint likelihoods to modulate selected GRPO advantages~\citep{2606.11709}. It discards uniform-outcome groups, and its token partitions remain heuristic rather than a demonstrated separation of style from reasoning. Distilled RL instead redistributes positive-advantage credit using teacher likelihoods~\citep{arxiv260717247}; in its comparisons, this sustains gains that the tested OPD and simple OPD--RL mixtures do not. Together, these constructions show how teacher evidence can refine reward optimization without supplying a distribution the student must continue matching.

SC-GRPO makes this distinction explicit through two branches~\citep{arxiv260618810}. With at least two correct responses and one failure, teacher--student divergence downweights locally aligned tokens while preserving reward-advantage signs. With all responses incorrect, it instead constructs a diversity pseudo-advantage from the divergence weights; the remaining group types use standard GRPO. Neither weighted branch adds a matching gradient. Its additive-distillation controls underperform GRPO on LiveCodeBench. This supports credit shaping in that setting while leaving open when direct matching is beneficial; teacher information and teacher imitation need not play the same role.

\textbf{Gating uses outcomes to decide which teacher updates survive.} RG-OPD retains trajectories according to the agreement between group-advantage sign and teacher--student sequence-likelihood ordering, then applies the same reverse-KL loss to retained responses~\citep{2607.04037}. OPDVR makes the decision at the token level, keeping teacher coefficients that reinforce correct responses or suppress incorrect ones; its group-relative variant uses advantage sign instead~\citep{2608.24696}. OPDVR's reversed-gate control performs worse despite comparable masking, supporting the direction rule over sparsity alone. The granularity matters: admitting a response preserves its internal matching signal, while admitting individual coefficients can remove local contradictions within that response. Group-advantage sign must also be distinguished from correctness: even with binary correctness rewards, zero group advantage can represent either uniformly correct or uniformly incorrect responses.

\textbf{Residual credit changes the signal rather than selecting it.} GC-OPD subtracts a normalized distillation score from normalized group reward and distributes the resulting discrepancy across tokens~\citep{2608.19181}. This addresses cases where teacher likelihood favors a lower-reward response: the disagreement becomes corrective information. RA defines a different residual, teacher minus student probability, and combines centered distributional guidance with outcome credit~\citep{arxiv261011519}. Its local squared-distance identity holds before response centering and clipping, not for the complete algorithm. In the accompanying Co-RA study, an adapting teacher improves student mean correctness while its own solver score falls, with mixed changes in student pass@$8$. Together these results motivate evaluating teacher guidance by its contribution to student learning, beyond standalone teacher accuracy.

Verification can also reshape the matching target or its weight before optimization. Graded property-equivalence scores weight forward-KL matching~\citep{2605.13501}. CriPO distinguishes criteria that no sampled response satisfies from criteria that some responses satisfy but aggregate reward fails to reinforce~\citep{2607.18082}. It uses criterion-conditioned matching for the former and a counterfactual teacher to locate and protect useful behavior in negative-advantage responses for the latter. SPOT instead probes possible continuations at branch points and uses verified outcomes to reweight a teacher prior~\citep{2608.04419}. Task evidence thus acts through loss weighting, support selection, or the target distribution itself.

When the task admits a useful description of progress, the teacher can supply process credit without dense policy matching. SPEAR rewards ordered overlap between symbolic reasoning milestones, while Cliff uses a verified teacher solution to locate the student's first mistake and reshape credit around that boundary~\citep{2608.26550,2609.02817}. These signals answer different questions: whether the attempt follows relevant milestones and where it first ceases to be valid. Their limitations follow from the same distinction. Symbolic overlap does not establish logical validity, and error-boundary credit depends on judge localization; Cliff also does not unconditionally reinforce every token before the judged error. The choice of process signal should therefore follow the evidence a task can reliably expose.

\subsection{Sequential and Alternating Learning}\label{subsec:sequential_training}

\textbf{Training schedules determine what distillation prepares and how its role changes during reward learning.} Preparation can have value that immediate accuracy does not reveal. PRISM inserts black-box adversarial pre-alignment between SFT and verifiable-reward RL, using perception and reasoning scores from a multimodal discriminator to reward student rollouts~\citep{2604.28123}. Its limited immediate accuracy gains are followed by improved GRPO performance on the evaluated 4B and 8B models. In its text setting, RL Starts before RL likewise finds a small initial accuracy gain but a larger advantage after subsequent reward learning~\citep{2609.28145}. PRISM uses a learned response-level proxy, whereas RL Starts before RL uses distribution matching. Both studies distinguish the behavior acquired during preparation from its value for later learning.

Evaluating preparation by later learning can reverse the ranking of matching objectives. In RL Starts before RL, reverse KL leads before RL whereas forward KL leads afterward on fourteen of twenty tasks when preparation uses student trajectories. Student-generated preparation also outperforms teacher-generated trajectories after RL under either divergence. In the framework of Section~\ref{subsec:support}, the objective determines which probability discrepancies are emphasized, while the rollout policy determines where they are corrected. Their value therefore depends on the intended learning sequence, making \emph{downstream learning potential} a distinct property of a distilled model. Associated entropy and alignment measurements do not establish that broader support or higher entropy causes the later gains. Nor do these studies establish fixed-budget superiority. PRISM matches the total number of policy-update steps but adds discriminator compute, while RL Starts before RL omits preparation cost from its comparison.

The preparation mechanism should address what limits the next learning phase. Teacher-side reward training followed by teacher-rollout warm-up and student-prefix OPD improves the behavior available for transfer~\citep{2605.12483}. Stronger-model demonstrations also prepare a useful teacher in that study, although the RL-trained teacher transfers more effectively in the comparison. Policy Reheater instead addresses a student policy that has become too concentrated to benefit from further RL~\citep{2606.00755}. It softens the current model's predictions through self-distillation before reward learning resumes. Its advantage over increasing sampling temperature is specific to the tested low-entropy regime, and already high-entropy models can destabilize. Improving the teacher and restoring student exploration thus address different obstacles to subsequent learning.

A second decision is how long guidance should remain active once reward learning begins. Gradual withdrawal can change either where matching is applied or how strongly it acts. STEPS concentrates privileged supervision in an early window and restores GRPO on annotated spans as distillation fades. SAF-OPD selects and bounds strong sampled-token signals before annealing their coefficient to zero~\citep{2605.10194,2607.29209}. The former reallocates span-level reward and matching updates, while the latter changes the strength and duration of a signal already combined with GRPO. Both motivate a finite teaching phase; SAF-OPD further isolates an annealing benefit in its Qwen3-4B mathematics ablation. Its smallest code model nevertheless remains below OPD alone in the reported comparison, so teacher retirement cannot be assumed to improve every setting.

Separating phases from the outset makes a stronger commitment than gradually withdrawing guidance. Sequential Beats Joint finds that reverse-KL OPD followed by verifier-reward GRPO outperforms joint alternatives on its logic tasks, with smaller or tied advantages in mathematics~\citep{2609.04108}. Its controls support interference between acquiring teacher-supported behavior and optimizing reward for the evaluated stronger frozen teachers. This motivates learning a useful repertoire before sharpening it toward outcomes, while the task variation limits generalization to every teacher and objective.

Student progress can inform when to end guidance or which prompts should still receive it. RetireOPD makes a permanent transition to GRPO once both alignment progress and relative task competence meet its criteria~\citep{2609.20784}. PIVOT instead increases the fraction of prompts assigned to GRPO according to a fixed cosine schedule, using teacher perplexity averaged over student rollouts to decide which prompts fill that fraction. Lower-perplexity prompts receive GRPO, while the remainder receive OPD~\citep{arxiv261011167}. This difference affects both responsiveness and cost. Permanent retirement can end teacher use, whereas repeated minibatch routing retains teacher evaluation even for examples assigned to RL. Both depend on proxies for further teaching value. RetireOPD's sampled alignment statistic does not directly measure gradient conflict, and evidence for PIVOT remains limited to two few-shot intent-classification tasks with matched update steps.

A recurrent alternative in CoPD interleaves specialist reward learning and transfer throughout training~\citep{2604.27083}, as developed in Section~\ref{subsec:mopd_evolution}. Fixed handoffs separate behavior acquisition from reward optimization, adaptive decisions use the student's progress to select a transition time or allocate guidance, and recurrent transfer incorporates improvements in the specialists. Their costs and effects on retained behavior depend on the full learning sequence, developed in Section~\ref{sec:dynamics}.

%% file: figures/chapter_maps/05_rl_integration.tex
\begin{figure}[!htb]
\centering
\includegraphics[width=\linewidth]{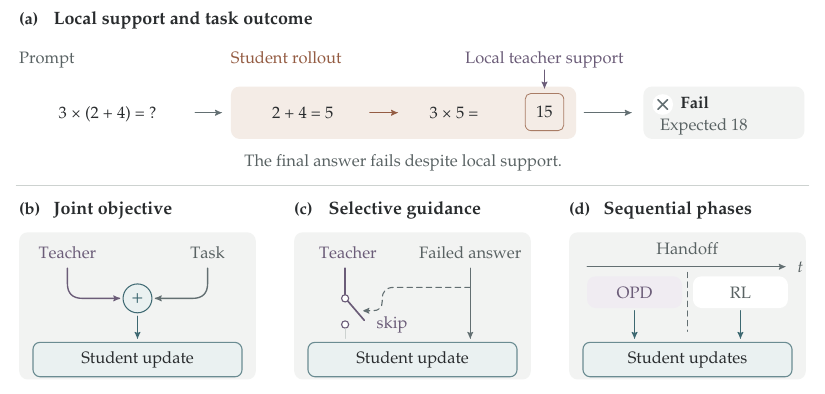}
\caption{Teacher preference and task success need not agree. In the illustrative rollout, the teacher supports a token even though the final answer is wrong. Joint objectives combine teacher guidance and task reward, the illustrated selective rule gates teacher guidance using the outcome, and sequential schedules assign the signals to different training phases. These choices can coexist in one method.}
\label{fig:rl_integration}
\end{figure}

%% file: chapters/06_mopd.tex
\section{Multi-Teacher OPD and Capability Integration}\label{sec:mopd}

\textbf{Integrating specialists requires deciding whose behavior to preserve at each student state.} A collection of strong mathematics, code, instruction-following, or multimodal policies does not itself provide one policy suitable for deployment across those capabilities. Multi-teacher on-policy distillation (MOPD) addresses this gap by querying specialist targets on trajectories generated by a shared student. It integrates behavior through policy supervision, whereas pooled reward training combines task rewards and parameter merging combines model weights. The result depends on which teacher can supervise a visited state, how much supervision each capability receives, and whether the specialists remain fixed while the student learns.

Figure~\ref{fig:teacher_routing} separates the check that admits a supervision source from the token allocation that determines how much each capability contributes to learning.

\input{figures/chapter_maps/06_teacher_routing}

\subsection{Teacher Eligibility, Routing and Target Composition}\label{subsec:teacher_composition}

Routing by domain is a natural starting point: MOPD trains specialists from a shared SFT model and asks the relevant frozen specialist to score each student trajectory~\citep{2606.30406}. Yet a teacher-replacement experiment shows sensitivity to teacher--student distribution mismatch. A clear task label therefore resolves only whose expertise is relevant, not whether that expertise transfers at the student's states. The study's sampled-token and mass-corrected top-$K$ variants also separate routing from the matching objective; correcting mass on retained tokens does not make vocabulary truncation an unbiased estimate of full-vocabulary KL.

Verified outcomes can make eligibility more specific than a domain label. MT-SDPO admits experts whose cached answers pass verification, then routes their critiques through a context-privileged self-teacher~\citep{2609.02548}. VG-OPD instead admits an expert only when its cached answer scores better on a criterion the student fails, and uses that expert's scores directly at selected token positions~\citep{2609.15404}. The distinction is both the unit of competence, a complete answer or an individual criterion, and the object transferred, a critique or local prediction scores. VG-OPD's scores form a detached auxiliary advantage alongside GRPO rather than an exact reverse-KL gradient. Its routing and localization controls favor the informed choices over random alternatives.

Verification establishes eligibility, not universal reliability downstream. MT-SDPO can fall below initialization when all teachers are admitted without checking, but verification still fails to improve the aggregate score, worst-domain score, or domain balance for one tested student~\citep{2609.02548}. Likewise, narrower differences between domain scores can reflect regression in an initially strong domain. A specialist must demonstrate relevant competence, and its contribution must still be evaluated through the student that receives it.

TrustMOPD provides a contrasting route when checked answers are unavailable~\citep{2609.23697}. It weights specialists by calibrated changes in their next-token preferences from a shared pre-RL reference, using each teacher's average change on a fixed sample for calibration. These detached reliability proxies enter a weighted reverse-KL update. Their interpretation depends on the specialists sharing that reference and concerns preference change rather than verified correctness. Table~\ref{tab:teacher_routing} contrasts routing critiques through a self-teacher with directly matching specialist predictions.

Once teachers are eligible, composition determines how their competing priorities enter a common target. SMOPD organizes specialists by reward priority and combines their guidance with a balanced anchor~\citep{2608.03092}. Another choice is whether to transfer complete policies or their improvements over a reference. $\Delta$-MOPD subtracts each teacher's initialization logits and adds the resulting shifts to the student's initialization logits~\citep{arxiv261010460}. TrustMOPD uses preference changes to weight supervision from teacher policies, whereas $\Delta$-MOPD transfers those changes as the target itself. For a teacher sharing the student's initialization, the shift construction reduces to matching that teacher's policy. Its paired mathematical experiments favor shift targets when composing three teachers, while the two target constructions perform comparably under interleaved routing to one teacher at a time. This distinction matters when specialists have different origins, though the reported norm-based balance diagnostics do not separate the magnitude of post-training change from differences in logit scale or confidence.

The divergence determines how composition treats disagreement among specialists. For the analyzed sequence-level objectives, forward KL yields an arithmetic teacher mixture and reverse KL a normalized geometric aggregate~\citep{arxiv260938666}. Arithmetic mixing preserves probability contributed by a specialist even when others assign little mass to the same response. Geometric aggregation instead penalizes such disagreement, connecting teacher composition to the probability-support distinction in Section~\ref{subsec:support}. Constructed examples show that a teacher assigning very low probability to useful responses can suppress them in the geometric aggregate even at small weight.

The level at which this composition is defined is essential. The sequence-level reverse-KL aggregate's next-token conditional includes a partition factor over possible continuations, so it is generally different from a locally normalized average of teacher log probabilities. These results characterize the specified sequence objectives, not empirical comparisons of trained language models or a universal identification of on-policy sampling with reverse KL. They explain why choosing how to combine teachers also requires choosing how their disagreement should affect the student's target.

\begin{table}[!htbp]
\centering
\small
\setlength{\tabcolsep}{3pt}
\renewcommand{\arraystretch}{1.08}
\begin{tabularx}{\linewidth}{@{}P{0.13\linewidth}P{0.24\linewidth}P{0.25\linewidth}>{\raggedright\arraybackslash}X@{}}
\toprule
\textbf{Method} & \textbf{Routing evidence} & \textbf{What is routed} & \textbf{Source of student update} \\
\midrule
\textbf{MOPD} & Prompt domain & One specialist per response & Specialist scores on student prefixes \\
\addlinespace
\textbf{MT-SDPO} & Verified expert answers & Expert critiques; correct peers supply separate anchors & Conditioned moving-average self-teacher distribution \\
\addlinespace
\textbf{VG-OPD} & Failed rubric criteria and verified expert answers & Eligible experts and sparse masks, at most one expert per token & Detached negative discrepancy advantage alongside GRPO \\
\addlinespace
\textbf{TrustMOPD} & Calibrated preference changes from a shared pre-RL reference & Weights over experts at each prefix & Weighted sum of expert-specific reverse-KL losses \\
\bottomrule
\end{tabularx}
\caption{Different objects routed in multi-teacher OPD. An expert can supply context for a self-teacher or provide the predictions used directly in the student update.}\label{tab:teacher_routing}
\end{table}

\subsection{Capability Balance and Retention}\label{subsec:mopd_allocation}

\textbf{Capability balance depends on both exposure to supervision and compatibility of the resulting updates.} A domain may receive too few supervised tokens, retain a large teacher gap that the student cannot readily close, or lose performance during updates for another domain. The methods below intervene at different points: sample selection, allocation across domains, and the optimizer's parameter displacement. These interventions address different sources of imbalance, so equal prompt counts or a small aggregate loss alone cannot establish successful consolidation.

A specialized student may need to recover general capabilities while preserving its specialty. CaMOPD alternates recovery and preservation, using different teacher--student gap scores to select samples for each goal~\citep{2605.27115}. Uncertainty-Calibrated MOPD instead filters candidate trajectories and token updates using specialist and general-teacher guidance~\citep{2608.26735}. Both make retention an explicit training concern. Their evidence also distinguishes retention from universal improvement: CaMOPD improves general-capability averages in its role-play and medical settings without uniformly improving the specialty.

A second imbalance arises because prompts contribute unequal amounts of supervision. Open-MOPD finds that instruction following receives roughly a fifth of prompts but only one percent of supervised tokens beside longer mathematics and code responses~\citep{2608.19098}. Token-share balancing corrects this exposure mismatch. D$^3$-MOPD asks a further question: is the remaining mismatch still learnable? It allocates prompts using both normalized teacher--student KL and recent progress, with a floor to retain every domain~\citep{2608.24987}. A large but stalled gap can thus receive less attention than a smaller, shrinking one. Exposure balancing and progress-based allocation address different bottlenecks; neither can be replaced by equal prompt counts. D$^3$-MOPD's reported saving is in rollout steps to a baseline score, rather than complete expert preparation and training cost.

A third imbalance concerns what an update does after data have been allocated. Even adequate domain exposure does not ensure that the optimizer's step preserves each capability. UP-MOPD addresses this possibility by constraining AdamW's proposed parameter displacement using current domain gradients~\citep{arxiv261008398}. Acting on displacement accounts for optimizer effects that raw-gradient conflict checks can miss. The constraint, however, describes first-order changes on the current batch rather than long-term freedom from forgetting. Evidence for conflict intervention is also mixed. In UP-MOPD's medical--general setting, the paired benchmark differences against M-OPD for one model evaluated late in training are not statistically significant after multiple-comparison correction, while Open-MOPD finds that its tested conflict masking and consensus targets hurt performance even when allocation changes help~\citep{2608.19098}. Together with the additional cost of computing domain gradients in UP-MOPD, these results make optimizer intervention a distinct tradeoff, not a default remedy for every capability imbalance.

Successful integration must be assessed by the capabilities retained, not only their average. A controlled comparison of multi-teacher OPD, parameter merging, and pooled reward training finds similar aggregate scores but much larger differences on individual benchmarks~\citep{2608.27409}. OPD tracks expert performance more consistently, while alternatives lose more instruction-following improvement. The experts' weight changes also exhibit different domain relationships, but this geometry does not identify a single cause of transfer or interference. Nor is a cheap fusion stage necessarily a cheap pipeline: OPD fusion costs less than one-fifth of per-domain RL in the comparison, whereas including prerequisite experts raises total cost to 1.14--1.19 times that baseline. Domain retention and teacher preparation therefore belong in the same assessment.

Distillation and merging can instead be successive stages of one design. USA prepares independently distilled language-agent experts for later merging, applying larger sharpness-training perturbations to parameters identified as more update-sensitive during warm-up~\citep{arxiv260934225}. Comparisons with a fixed merging operator favor this preparation, while randomized or inverted assignments weaken transfer. The contribution is to train experts for their eventual combination, rather than assuming that individually strong policies will also merge well.

\subsection{Integration Pipelines and Evolving Specialists}\label{subsec:mopd_evolution}

Routing and allocation assume that useful specialists are available. A complete pipeline must also establish the student that receives their guidance, construct the specialist pool, and decide whether that pool stays fixed during transfer.

Preparation determines both the student states available for refinement and the granularity of the teachers. Two-phase agentic distillation first pools expert traces for initialization, then queries fixed task experts on the shared student's trajectories, recovering much of the specialization lost during pooling~\citep{2606.30044}. Socio-Foundation instead first consolidates task experts into capability teachers on fixed histories before integrating them on student-generated histories~\citep{arxiv261008967}. The former prepares the recipient; the latter also restructures what is transferred. Socio-Foundation's comparisons favor the hierarchy while still exposing tradeoffs between outcome attainment and other simulated behaviors.

Specialization need not follow broad academic domains. Platform-specific GUI teachers and category-specific software-engineering experts provide useful divisions within an environment family~\citep{2607.04425,2609.23377}. In the software-engineering study, integration recovers most or all of the specialists' gains over the common base across three categories, although differences in training pools, budgets, and checkpoint selection limit comparison with joint RL. These cases make teacher granularity a substantive design choice: the specialist should capture a recurring source of behavioral variation that one general policy must eventually handle.

System reports combine these preparation and integration choices in different training schedules~\citep{2601.02780,2603.19220,deepseekv4}. MiMo-V2-Flash combines specialist log-ratio guidance with outcome advantages and compares the joint update with reward-only and distillation-only variants on mathematics and code. Nemotron-Cascade 2 instead reuses intermediate models with strong domain-specific performance from its own training pipeline, including the initial SFT model, to rebalance capabilities after sequential RL. Its sampled-token log-ratio updates and stage comparisons provide a concrete checkpoint-reuse design, without establishing equal-compute superiority for the complete pipeline. DeepSeek-V4's implementation of dense multi-teacher supervision is discussed in Section~\ref{subsec:systems}.

Fixed-specialist pipelines transfer an existing repertoire; evolving specialists can supply new behavior during integration. CoPD interleaves capability-specific reward training with bidirectional policy-target exchange on each branch's own rollouts~\citep{2604.27083}. CoDistill-GRPO couples models asymmetrically: the small model receives a scalar distillation reward from the large one, while the large model reuses small-model trajectories with importance reweighting~\citep{2605.08873}. Thus, joint evolution can mean exchanging targets in both directions or combining target flow one way with trajectory flow the other. Their evaluated recipes support these forms of coupling, but do not establish a general advantage for every component or an end-to-end cost benefit. The extra branches remain part of the training system even when deployment uses one consolidated student.

Caching frozen specialists and refreshing evolving branches impose different requirements for running teachers, even when both pipelines deploy a single student. Teacher placement and target caching therefore depend on the chosen teacher roles and update schedule, as developed in Section~\ref{subsec:systems}. When student actions also alter the environment, teacher selection must work alongside history reconstruction and the use of delayed outcomes. These interaction-specific dependencies arise whether supervision comes from one teacher or a specialist pool.

%% file: figures/chapter_maps/06_teacher_routing.tex
\begin{figure}[!htb]
\centering
\includegraphics[width=\linewidth]{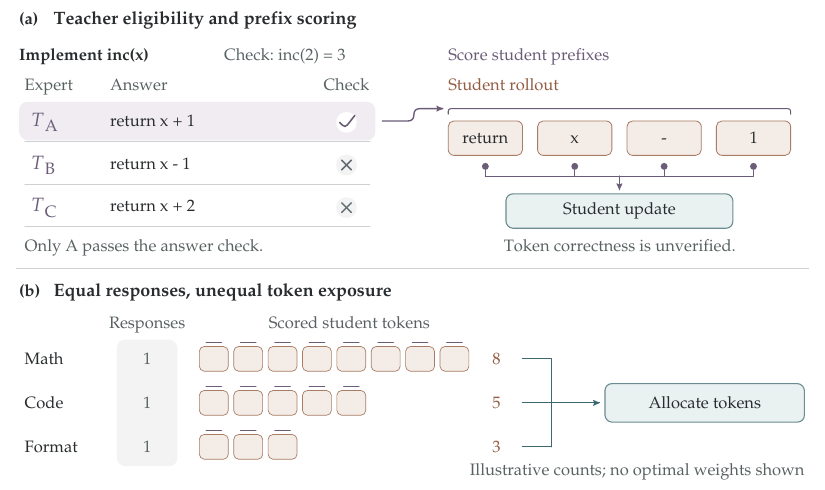}
\caption{Teacher eligibility and supervision allocation. In the upper example, an expert with a successful answer supplies scores at prefixes of the student's failed response; the answer check leaves token-level reliability unresolved. The lower example illustrates how equal response counts can yield unequal token exposure across capabilities, motivating a separate allocation rule.}
\label{fig:teacher_routing}
\end{figure}

%% file: chapters/07_agentic.tex
\section{Agentic OPD}
\label{sec:agentic}

\textbf{For an agent, on-policy supervision must follow both the generated action and the state that produced it.} Each action can trigger a tool response or environment transition, and memory can rewrite the history visible at the next invocation. A teacher may therefore score the student's token under the wrong state even when that token was sampled on policy. The mechanisms in Sections~\ref{sec:signal}--\ref{sec:mopd} remain available, but now interact with state reconstruction, cross-turn credit, execution control, and rollout depth. These are consequences of an interaction loop, rather than response length alone.

Figure~\ref{fig:agent_states} follows one tool interaction to distinguish the history seen by the acting student, the tokens that carry its loss, and additional information deliberately supplied to the teacher.

\input{figures/chapter_maps/07_agent_states}

\subsection{State Identity and the Unit of Supervision}
\label{subsec:interaction_units}

An interaction turn can contain several decisions, while environment observations condition subsequent actions without themselves being generated by the student. Tool-use and retrieval methods therefore retain observations in the scoring history while applying the distillation loss only to model-generated tokens~\citep{2605.07725,2608.24310}. Masking retrieved passages, for example, preserves their role in conditioning the next query. \textbf{The unit used to allocate supervision need not be the unit that receives the gradient.} SOD uses sampled discrepancy to weight tool-use steps; its pure-distillation control favors step weighting over uniform OPD in the tested Python-tool setting. OPDSearch+ further separates this masked distillation stage from later teacher-free RL, illustrating that the loss mask and the training schedule are independent choices.

The teacher's information and identity are separate from this loss mask. Skill-SD and MAD-OPD condition token-level guidance on skills and multi-agent debate, respectively~\citep{2604.10674,2605.01347}. UI-MOPD adds platform-specialist routing with reward-conditioned masking~\citep{2607.04425}. Privileged skills and debate change \emph{what information the supervisors see}; MAD-OPD also combines several teachers, while UI-MOPD routes among platform specialists. Transferring privileged information also differs from providing it directly at deployment. Skill-SD's direct skill-prompting control underperforms plain GRPO on held-out AppWorld and Sokoban tasks~\citep{2604.10674}.

\subsection{State Reconstruction, Replay, and World Models}
\label{subsec:state_alignment}

State reconstruction must distinguish deliberate teacher privilege from an accidental change to the student's history. In conversations with incrementally revealed requirements, MAIGO supervises intermediate turns with cleaned histories and also supervises the final answer after all requirements arrive, whereas Canonical-Context OPD matches only that final answer~\citep{2605.27186,2605.30251}. Both can use the complete task for final-answer supervision; they differ in whether intermediate responses also receive direct guidance. A world-model teacher can instead receive an observed transition unavailable when the student made its prediction, as in CoMAP~\citep{2606.02372}. Additional observations or a cleaner presentation can change the supervisor's judgment; it need not rewrite the state assigned to the student's original action.

Choosing a training history is a separate decision. Prefix Replay samples fresh student actions from stored teacher histories, reducing interaction cost~\citep{2607.04763}. TAP uses fixed teacher contexts after structural pruning to keep a weakened student out of unproductive trajectories~\citep{arxiv261009074}. Both provide local on-policy action supervision on deliberately chosen histories, rather than complete coverage of free student execution. Their advantages are setting dependent: Prefix Replay essentially ties ordinary OPD on its reported search comparison, and TAP's recovery advantage over full-trajectory OPD disappears on WebShop at 80\% feed-forward-network pruning. A history can be useful for learning without being representative of the states the deployed student will reach.

Once an action has been collected, however, its conditioning history must be reproduced faithfully. MemOPD shows that reconstructing a persistent conversation can reintroduce information a memory-compressing agent had already removed~\citep{2608.07068}. Scoring the action under that altered history corrupts likelihoods and can spuriously activate policy-ratio clipping. Preserving visible content, token positions, and action masks is therefore part of on-policy supervision, not merely bookkeeping. Teacher-only hindsight remains compatible with this requirement as long as it changes the teaching view without being silently inserted into the student's view.

Student participation in state generation also does not determine the learning objective. QUEEN uses current-model chess proposals inside engine-guided search, then consolidates analyses into targets for subsequent SFT~\citep{arxiv261003695}. This is search followed by imitation, rather than conditional distribution matching at the original action prefixes. The distinction prevents improvements from a broader search-and-training pipeline from being attributed to local OPD alone.

\subsection{Cross-Turn Credit Assignment}
\label{subsec:turn_credit}

Once states and action positions are aligned, the next issue is what later evidence says about an earlier decision. OpenClaw-RL turns observed consequences into rewards and selected hints for a privileged self-teacher~\citep{2603.10165}. Gold actions and reusable skills offer other sources of guidance~\citep{2605.29584,2605.15155}. Privilege need not fix the direction of teaching: UCOB compares skill-conditioned and unconditioned rollouts and lets the higher-return view teach the other at a repeated state~\citep{2606.29502}. The design choice is whether a view is presumed useful because of its information or admitted because its behavior succeeds. OpenClaw-RL's hint-selection comparison also shows that random hint selection can destabilize learning.

Terminal feedback arrives after several actions, motivating hindsight guidance at selected action spans, search-query positions, or usable turns~\citep{2605.17873,2605.18299,2606.11559}. The shared idea is to localize what later evidence can teach about earlier decisions. HINT-SD, for example, generates corrective feedback from a failed trajectory and uses a feedback-conditioned self-teacher for reverse-KL matching on selected action spans, retaining each action's original history. Such localization remains heuristic when several turns share an outcome or query quality is inferred from thresholded answer overlap.

Hindsight can change either a target or the credit attached to an action. SERL uses hindsight-conditioned likelihood gaps to positively rescale executable-action GRPO updates, preserving their reward-derived direction, and adds a separate action-only forward-KL term~\citep{2605.19447}. MIMESIS instead applies positive coached likelihood weights to existing agent tokens~\citep{arxiv261009484}. A small weight there weakens reinforcement without directly penalizing a disfavored token. The same description, ``teacher weighting,'' therefore conceals different effects depending on whether the weight multiplies a signed reward advantage or a likelihood objective.

These choices also separate hindsight from temporal credit. A later outcome can improve a local target at the original history, as in HINT-SD, without computing a return over subsequent teacher--student discrepancies~\citep{2605.17873}. SERL additionally modulates outcome-derived credit, but its separate matching term means that reward sign does not determine the complete combined update. Which information the teacher sees and which consequences enter an action's gradient remain distinct questions (Section~\ref{subsec:temporal_credit}).

Multi-agent interaction adds responsibility across roles to credit across turns. MAS-OPD combines ordinary teacher--student log-probability guidance with a contrast between intended and alternative roles, while privileged teacher context contains verifier judgments about completed earlier turns~\citep{arxiv260934234}. The role contrast encourages specialization; the retrospective context guides whether to revise or retain a position after conflict. Their reported effects are complementary, but each role still learns from its own generated tokens. The signal retains the student log-probability term and is not an exact joint-trajectory KL gradient. Role-specific supervision therefore extends coordination without collapsing all interacting policies into one ordinary sequence model.

Allocating guidance across turns remains a separate choice. Healthcare AI GYM's turn-level TT-OPD combines outcome-conditioned self-teaching with GRPO and length-sensitive reward~\citep{2605.02943}. ATOD shifts weight toward GRPO over training and attenuates turns with weak discrepancy and surprisal signals; PCSD aggregates local teacher gaps to stabilize selection~\citep{2606.27814,2608.01837}. These methods regulate where and how strongly guidance acts. Their proxies do not by themselves locate a causal error boundary, even when they improve the allocation of a fixed supervision budget.

\subsection{Teacher Intervention, Handoffs, and Recovery}
\label{subsec:intervention}

\textbf{Teacher intervention changes which states the student will encounter.} DAgger permits expert execution through a mixture policy~\citep{1011.0686}; agentic handoffs make takeover conditional on the unfolding interaction. DASH-OPD accumulates discrepancy across turns and uses different entry and exit thresholds to avoid reacting to every transient difference~\citep{2607.29078}. STI-OPD instead makes takeover stochastic, increasing its probability with discrepancy~\citep{arxiv261010878}. Both regulate when to intervene, but disagreement can still reflect a benign alternative or miss a shared error.

Testing the consequence of an intervention addresses that limitation more directly. FTB compares paired student continuations after a candidate teacher bridge and keeps only accepted bridges~\citep{2608.01953}. Its motivating study finds that some high-disagreement interventions worsen outcomes, demonstrating why detecting mismatch and validating recovery are different problems. The validation still uses a teacher-preference proxy, and extra continuations cost computation. Intervention also changes the training distribution: STI-OPD's conditional importance weighting does not restore histories altered by teacher actions or turn its clipped update into an exact trajectory-level correction. A useful bridge must therefore be judged by the subsequent behavior it enables, with its sampling effects kept explicit.

Ending a local intervention and ending teacher supervision operate on different timescales. DASH-OPD controls execution handoffs, while FTB tests and retains teacher bridges within training trajectories. RetireOPD instead uses alignment progress and relative task competence to remove the teacher from subsequent training, as detailed in Section~\ref{subsec:sequential_training}~\citep{2609.20784}. A handoff changes the ensuing interaction, whereas retirement removes teacher guidance from the remaining training.

\subsection{Continuation Depth and Applicability}
\label{subsec:agent_horizon}

Shortening an interaction saves work only if the omitted states are not essential to learning. TCOD expands the student-controlled horizon through a curriculum, whereas TurnOPD combines KL concentration with the observed completion depths of successful episodes~\citep{2604.24005,2607.05804}. The latter's success-based lower bound prevents a small local discrepancy from ending every rollout before useful completion states. Its ALFWorld comparison improves time and score, though periodic full-depth probes add cost. On multi-hop tasks, vanilla OPD scores slightly higher at equal update steps, while TurnOPD scores higher at equal training time. Horizon selection thus needs evidence about task completion in addition to local agreement.

Stopping also leaves a choice about where to resume. STRIDE couples a teacher-likelihood stopping rule with restarts from selected retained prefixes~\citep{2609.14636}. Restarting can expose continuations that truncation would otherwise omit, but it requires useful prefixes: removing quality filtering can collapse effective training length. Its accuracy--time tradeoff varies by domain and student size. Together, curriculum, completion-aware stopping, and selective restarts address how to reach useful later states without repeatedly paying for every full interaction.

Intervention and horizon control consequently act on more than the current loss. A teacher bridge changes the next state, a stopping rule removes later states, and a restart returns training to a selected history. Agentic efficiency depends on which learning opportunities those decisions create or remove, connecting execution control to the broader training dynamics examined next.

%% file: figures/chapter_maps/07_agent_states.tex
\begin{figure}[!htb]
\centering
\includegraphics[width=\linewidth]{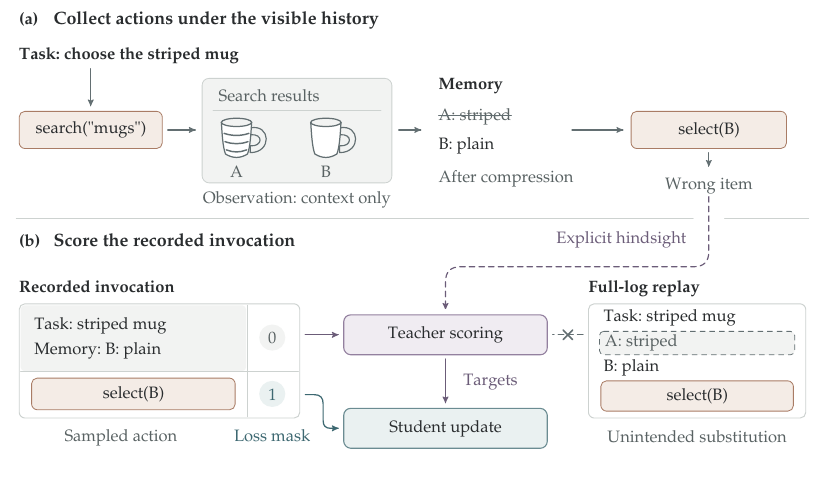}
\caption{Agentic supervision depends on the action and its invocation state. Tool observations provide context, while the loss applies to student actions. Here, memory compression removes information before an action is sampled. The crossed replay route restores that information and therefore scores the action under a different history. Deliberate teacher-only hindsight enters separately from the history visible to the acting student.}
\label{fig:agent_states}
\end{figure}

%% file: chapters/08_training.tex
\section{Training Dynamics and Efficiency}
\label{sec:dynamics}

Useful supervision has to be delivered repeatedly while the student changes. The capacity and budget tradeoffs in Section~\ref{subsec:distillation_scaling} therefore depend on which experience training produces. Shorter rollouts omit later states and replay reuses earlier states, changing $d_\mu$ in Equation~\ref{eq:common_match}. Selecting positions changes their weight $w(s)$, whereas restricting vocabulary items changes the support of supervision at a given state. These efficiency choices change both computation and the information available for learning.

Figure~\ref{fig:training_budget} contrasts computation already performed before masking, states omitted by early stopping, and additional updates obtained by revisiting a cached batch.

\input{figures/chapter_maps/08_training_budget}

\subsection{Prompt Breadth and State Coverage}
\label{subsec:coverage}

Prompt count is an incomplete measure of the experience available to an on-policy learner. Repeated generation from a changing student can expose new prefixes even when the prompt stays fixed. One-query mathematical training illustrates this possibility: repeated trajectories approach full-corpus performance, and a separate representation analysis finds broad, though incomplete, state coverage~\citep{2609.04172}. Selecting harder prompts can also help, but matched-truncation controls substantially weaken their advantage~\citep{2609.05198}. Some apparent benefit of difficulty therefore comes from longer supervised trajectories, without establishing whether later states or simply additional training tokens are responsible.

The value of input diversity also depends on the initial student and subsequent renewal. Small teacher-generated question sets can match real-data training in some model pairs, while removing explicit questions has sharply different effects across pairs~\citep{2609.14193}. In another study, expanding a prompt bank hurts when initial-student responses are frozen but helps when responses are regenerated~\citep{2609.25048}. Unequal token consumption and single-run configurations limit that comparison, yet the reversal cautions against treating prompt breadth independently of the rollout policy.

These findings distinguish three aspects of coverage: breadth across inputs, depth along trajectories, and renewal as the student changes. A small prompt set can still produce diverse, evolving states; a large fixed response bank cannot provide fresh student behavior. ``Data-free'' question acquisition likewise leaves student generation and teacher scoring as essential sources of training experience.

\subsection{Supervision Selection and Trajectory Control}
\label{subsec:selection}

Selection decides where to spend supervision. It can act before generation by choosing prompts, after generation by retaining or weighting complete responses, or within a response by choosing segments or individual positions. At a selected position, it can also restrict which vocabulary items enter the loss.

\subsubsection{Allocating prompts and complete responses}

A useful prompt must expose a learnable gap and admit reliable guidance. PACED emphasizes problems at intermediate student competence, whereas Transfer-Stratified OPD uses joint teacher--student success estimates to select forward-KL acquisition, reverse-KL correction, or no update~\citep{2603.11178,arxiv261005974}. The latter uses forward KL when only the teacher succeeds, reverse KL when both succeed but the teacher remains stronger, and skips prompts without an estimated teacher advantage. Allocation thus depends on both student competence and available guidance. Small initial samples can misclassify competence, and zero observed success does not rule out useful token-level supervision.

Student difficulty alone leaves the teacher's contribution unspecified. Uni-OPD balances difficulty and response correctness while calibrating teacher--student returns using outcomes; DeltaPrompts seeks teacher--student answer disagreement together with teacher consistency~\citep{2605.03677,2605.15532}. These approaches distinguish a struggling student from a teacher that offers a reliable alternative. Competence, disagreement, and reliability can inform the same allocation rule without becoming interchangeable signals.

At the response level, compatibility and task success offer complementary evidence. Lower teacher perplexity on erroneous responses predicts recovery from truncated prefixes in SCOPE, whereas RA-OPD filters trajectories when verifier outcomes conflict with the sign of the average distillation return~\citep{2604.10688,2608.27960}. The former estimates whether a trajectory remains recoverable; the latter rejects a mismatch between outcome and aggregate update signal. Neither locates the erroneous tokens. FiRe-OPD combines hard trajectory filtering with soft token weighting, illustrating why a response-level eligibility decision can coexist with finer allocation inside the retained response~\citep{2606.02684}.

\subsubsection{Locating useful supervision within a response}

Local selection asks which parts of a response carry useful supervision. Continuation probes estimate whether an intermediate prefix still permits a correct answer; R2-OPD uses these estimates to mask segments whose process and distillation rankings disagree~\citep{2608.19408}. This supplies task-related evidence at the cost of extra sampling and estimation variance. More aggressive masking can sharply hurt performance, so identifying low-value segments does not make retaining less supervision uniformly beneficial.

Distribution scores avoid those probes but require a choice of what to favor. SelecTKD emphasizes student proposals accepted by teacher verification, while Teachability-Aware OPD favors disagreement supported by teacher mass on the student's candidates~\citep{2510.24021,2605.26844}. Compatibility protects usable behavior; supported disagreement identifies a possible learning opportunity. TIP further distinguishes confident from uncertain disagreement, and DIAL-OPD discounts large log-ratios arising where both models assign negligible token probability~\citep{2604.14084,arxiv261011659}. These refinements qualify the meaning of a discrepancy. None makes distributional disagreement a direct test of token correctness or ultimate reachability through training.

A different criterion is whether the teacher's additional training or information changes its advice. S2D-OPD selects states where RL most changes the teacher relative to a reference, while Visual-Advantage OPD weights supervision by the teacher's sensitivity to fine visual information~\citep{2609.29142,2605.21924}. Their selection and masking comparisons support concentrating supervision on informative differences. The interpretations nevertheless differ: a teacher's RL-induced change can be harmful, and visual sensitivity measures information dependence rather than correctness.

Estimating what an update will achieve brings allocation closer to learning benefit, but can be much more expensive than using properties of the trace. At the inexpensive end, PW-OPSD gives later positions more weight while retaining earlier supervision; its supporting diagnostics concern ambiguous positions on initially correct traces~\citep{2605.21606}. MetaOPD instead learns weights through the effect of a one-step update on reference solutions~\citep{arxiv261011989}. This estimates an outcome of learning rather than relying on position as a proxy, but nearly doubles measured training GPU-hours in its reported Qwen3-1.7B setting. The practical tradeoff is between the reach of a cheap structural assumption and the cost of estimating which updates will help.

Choosing a position still leaves the support of its target unspecified. SafeSteer~\citep{2606.02530} selects safety-relevant vocabulary items within a position. It changes which candidate tokens contribute to the loss, a distinct choice from deciding which states or sampled actions receive supervision.

Selection changes which parts of a generated trajectory contribute to learning. A mask applied after complete generation and teacher scoring can reduce subsequent work, but leaves those two costs intact. Earlier intervention can change the trajectory itself.

\subsubsection{Trajectory repair and behavior intervention}

Repair changes the behavior distribution itself. The intervention can respond to broad divergence from the teacher, as in AdaSwitch, or to a specific missed opportunity for reflection, as in Relay-OPD~\citep{2510.07842,2607.26057}. Both insert teacher-generated continuations into the student's trajectory. They therefore incur online teacher decoding and train on mixed histories rather than merely reweighting the original response.

Where the correction enters determines what remains of the student's mistake. BRTS adds a teacher response selected for correctness and student compatibility, although its fallback can still be incorrect~\citep{2605.09725}. MOTAB instead backtracks to an earlier student prefix and stitches a teacher continuation after the deviation through a revision marker~\citep{2605.19433}. Whole-response supplementation supplies an alternative route; backtracking preserves an explicit relation between deviation and revision. Teacher compatibility motivates both constructions, but does not by itself certify a response or repair point as logically valid.

Compression changes the supervised prefixes more extensively. MPD applies token-level KL to teacher-shortened versions of current-student reasoning traces~\citep{2605.08776}. Comparisons of trajectory sources and compressors find useful accuracy--length tradeoffs for this mixed-policy recipe. Although the student supplies the original trace, matching takes place on rewritten prefixes.

Student compatibility can also be built into a fixed training corpus. In Interleaved-Policy Distillation (IPD), a frozen teacher proposes tokens, a frozen initial student replaces poorly supported proposals, and the teacher continues from the revised prefix~\citep{arxiv250922230}. The resulting mixed trajectories become targets for offline SFT. Finetuning with Sampling likewise constructs model-matched traces before supervised training, using offline expert-conditioned resampling under the frozen base model's likelihood rather than scoring the evolving student's rollouts~\citep{arxiv261002140}. Both report settings in which tailored offline traces outperform the compared on-policy alternative. The benefit depends on the model receiving the traces, and IPD's teacher-propose advantage over student-propose generation shows that compatibility alone does not explain trajectory quality. These results position trajectory construction as both an alternative to OPD and a possible source of initialization for it.

\subsubsection{Horizon control}

Horizon control changes how far the student generates before learning from a trajectory. FOPD~\citep{2602.15260} begins with short student prefixes and expands their length during training. For Qwen3-8B-Base, the reported schedule takes $53.0$ hours on an eight-A100 virtual machine, compared with $148.3$ hours for full OPD, while benchmark accuracy remains within $0.5$ percentage points. These results exclude evaluation and depend on the selected student. Very short prefixes hurt out-of-domain accuracy for a weaker model, exposing the cost of withholding direct supervision from later states.

The central horizon tradeoff is whether the avoided suffix contains states that need their own training. Fixed early-window supervision can outperform post-generation token selectors while also avoiding suffix generation~\citep{2605.27028}. Comparisons of fixed and progressively extended horizons find useful accuracy--time tradeoffs in language models, but an autoregressive navigation control fails under early-only supervision when later states require separate learning~\citep{2605.31490}. The success of a short prefix therefore depends on how much its updates generalize beyond that prefix.

Adaptive horizons use evidence from the teacher to decide when further generation is worthwhile. Prune-OPD shortens subsequent budgets after cumulative compatibility failures; KAT stops when local teacher--student KL remains small~\citep{2605.07804,2606.09471}. These rules respond to different low-value suffixes: unreliable guidance versus vanishing discrepancy, which in KAT can reflect degenerate agreement at corrupted prefixes. Their benefit also depends on the overhead of checking the stopping condition. KAT evaluates random stopping and fixed prefixes, whereas Prune-OPD's advantage over equal-retention random stopping remains unestablished.

Training feedback can also determine how the horizon changes~\citep{2607.13124,2607.29494}. ShortOPD monitors repetition, truncation and effective response length when recovering pruned models, shrinking unproductive early rollouts and restoring longer generation as recovery proceeds. Adaptive FastOPD expands when boundary learning signals plateau and sufficiently many rollouts reach that boundary. Its more conservative multi-signal rule improves final accuracy at additional cost relative to a simpler overlap-based rule. These designs separate the potential usefulness of later states from the evidence needed to justify extending the budget to reach them.

The checking policy is itself part of this tradeoff. Flash-OPD adjusts verification intervals using accumulated low-overlap events and reuses the resulting teacher scores for training~\citep{arxiv261006105}. Repeated verification and prompt processing make its time savings smaller than its token savings, and one tested pairing incurs a small accuracy cost. An effective stopping rule must therefore save more work than it introduces while preserving useful later-state supervision.

\subsection{Refresh, Replay, and Staleness}
\label{subsec:freshness}

\textbf{Refreshing a score does not refresh a state.} A stored trajectory may come from an older student, carry student probabilities and advantages computed before recent updates, or retain scores from an earlier teacher. Exact reuse of a teacher score therefore does not preserve on-policy state collection.

A fixed cache can remain useful even after its generating policy becomes outdated. Lightning-OPD precomputes teacher scores on initial-student responses, removing live teacher inference from optimization; its cost accounting includes collection and precomputation, while its approximation depends on teacher consistency and additional conditions~\citep{2604.13010}. Semi-OPD retains the same states and teacher scores but recomputes student probabilities and advantages~\citep{arxiv261011291}. It outperforms fresh-rollout OPD in most of its mathematical pairings, and complete cached trajectories outperform their first quarter at a matched token budget. These results show that useful stored supervision can outweigh the value of renewal in some regimes, without making fixed states equivalent to current-student states.

The content of a cache matters alongside its age: a weaker policy can expose errors that complement the current student's attempts. Negative-Policy OPD exploits this possibility by mixing complete trajectories from a fixed weaker policy with current-student trajectories under the same teacher--student supervision~\citep{arxiv261007874}. Its reported gains span mathematical, coding, and scientific reasoning, with a fully cached endpoint that removes current-student generation. The weaker policy supplies states on which to learn; the teacher still determines which tokens to encourage or suppress. Thus, the benefit concerns the experience available in the cache, while pre-generation also shifts part of its cost outside online optimization.

Asynchronous training poses a different problem: it aims to follow the changing student but delivers trajectories late. In a coding-agent comparison, f-OPD finds a substantial loss from asynchronous delay and recovers most of it through freshness-aware control~\citep{2605.17862}. Near-Policy combines periodic rollout-policy refresh with a teacher--student differential-difficulty filter, although this heuristic does not measure trajectory age or ensure current-state coverage~\citep{2605.05940}. These controls address delayed tracking, whereas a fixed cache deliberately trains on a stable state distribution.

The update rule also changes tolerance to delay. In the AsyncOPD study, a prior-work sparse top-$K$ forward-KL implementation degrades less under stale rollouts than the study's PPO-style reverse-KL pipeline, despite the latter performing better with fresh data~\citep{2606.24143}. Recomputing advantages and applying a behavior-to-learner importance ratio corrects sampled actions at cached states; it cannot renew the states themselves. AsyncOPD's throughput gains concern the measured sparse asynchronous pipeline, leaving dense full-vocabulary and multi-node settings unresolved.

Inner-loop reuse spends additional student updates on supervision already acquired. Open-MOPD refreshes student-dependent candidate advantages while retaining teacher log-probabilities, and CLOOPD reports gains from extra student passes without another teacher pass~\citep{2608.19098,2609.24141}. CLOOPD's adaptive allocator does not outperform random allocation at the same reuse cap, leaving additional student updates as the better-supported explanation for the reported gains.

Refresh, replay, and reuse allocate computation to different parts of the learning process. Refresh acquires new states, replay draws further supervision from stored ones, and inner-loop reuse spends additional student updates on an acquired batch. Their value depends on how much useful supervision remains in stored states relative to the cost of reaching new ones. Refreshing teacher scores or student-dependent coefficients can improve an update without changing that state-coverage tradeoff.

\subsection{Initialization and Training Lifecycle}
\label{subsec:lifecycle}

\textbf{The student must be ready to benefit from the states and targets a schedule exposes.} Teacher preparation and self-teacher evolution change the available guidance (Sections~\ref{subsec:teacher_validity} and~\ref{subsec:self_evolution}), but cannot by themselves ensure that the student reaches useful prefixes or can fit the resulting targets. Initialization, guided exploration, and objective scheduling address these different obstacles. Their role also depends on whether OPD is the final training stage or preparation for task-reward learning (Section~\ref{subsec:sequential_training}).

Initialization affects whether the student reaches states on which teacher guidance is useful. An off-policy or supervised stage can establish compatible reasoning patterns before student-generated training, a role shared by mathematical cold-start experiments and the Qwen3 smaller-model distillation pipeline~\citep{2604.13016,2505.09388}. DeepSeek-R1's distilled students provide a contrasting pipeline in which teacher-trace training alone produces strong students~\citep{2501.12948}. Supervised preparation can thus support later OPD without implying that an on-policy stage is always necessary.

A transition into OPD can prepare the student or alter the histories it first explores. BIRD fits concise verified traces before self-distillation, while HPD uses fixed offline prefixes with limited student alternatives to prepare later rollout-based learning~\citep{2607.15736,2604.20244}. These methods change the initial policy; HPD's isolated student-sampled action does not make its fixed-prefix stage on-policy. Trust-Region Behavior Blending instead introduces teacher guidance into the sampling policy and anneals toward student-only generation~\citep{2605.31159}. It changes experience during the transition and requires online teacher decoding. Thus, preparing a compatible starting policy and guiding its first trajectories address related difficulties through different interventions.

For a pruned student, preparation also determines which damage OPD must repair. ReCal uses teacher-response prefixes and a pruning probe to emphasize vulnerable tokens when choosing channels, before a shared SFT--OPD recovery pipeline~\citep{arxiv261011332}. Benefits already appear after SFT, so the final improvement cannot be attributed wholly to more effective OPD. Moreover, the reported AIME gains accompany consistent IFBench declines. Preparing a student for successful reasoning transfer can therefore preserve a different set of capabilities from those needed for instruction following.

The objective can follow a separate schedule. CADENCE uses student coverage of the teacher's top-$K$ support to accelerate a transition from a mode-covering surrogate toward reverse KL~\citep{2607.16955}. Coverage feedback improves the tested mathematical recipe, but accelerates the schedule only after a threshold is crossed; it cannot by itself stop an aggressive baseline schedule from sharpening too early. Objective scheduling consequently governs how the student fits a target, independently of when new states are collected.

Successful preparation need not produce the best immediate distillation score. As Section~\ref{subsec:sequential_training} shows, the preferred distillation endpoint can differ from the best starting point for later RL. A schedule therefore has to connect what the student can learn now with what it must retain or learn next; optimizing each stage in isolation can favor a different policy from optimizing the intended learning sequence.

\subsection{Continual Adaptation and Retention}
\label{subsec:retention}

Repeated adaptation adds a retention objective to immediate transfer. Demonstration-conditioned SDFT accumulates skills across tool use, science question answering, and medical adaptation more successfully than the compared SFT pipeline, although some earlier capabilities still degrade~\citep{2601.19897}. Continual fact learning gives a complementary result: OPD retains earlier writes better than distillation on fixed teacher continuations, but varied-example SFT retains more~\citep{arxiv260711020}. Together, these comparisons suggest that renewed trajectories can help preserve learning across updates, while ordinary supervised diversity can provide a strong alternative. The relevant property is what training experience supports successive changes, rather than the on-policy label alone.

\subsection{Full-Pipeline Compute and Systems}
\label{subsec:systems}

The relative costs of preparing the teacher, generating student trajectories, and applying supervision determine which efficiency strategy is useful. A pretrained teacher may already be available, while a new RL-trained specialist adds preparation cost. Caching moves teacher scoring before student optimization, and batch reuse amortizes it over additional student updates.

These costs also change as the student learns. Comparisons with GRPO depend on how much teacher scoring and rollout generation each method requires~\citep{2604.10688}. Selective correction can reduce the teacher workload when fewer failed attempts remain eligible, as in SRPO~\citep{2604.02288}. A fixed cost per update can therefore obscure how the balance between generation and supervision changes over training.

Autoregressive generation and parallel scoring of a known sequence impose different bottlenecks. FOPD reduces both rollout and training work by stopping at a prefix, whereas SafeSteer restricts vocabulary support after the required states have been produced and may still score much of the teacher distribution~\citep{2602.15260,2606.02530}. Sparse losses therefore need not yield the same savings as shorter trajectories.

The value of teacher computation depends on what training it displaces. Against task-reward RL, Qwen3 reports $1{,}800$ GPU-hours for its 8B on-policy distillation stage versus $17{,}920$ for the compared RL stage, with a higher AIME24 result for distillation~\citep{2505.09388}. In pretraining, however, the alternative is to spend those resources on more student updates. MiniPLM compares online MiniLLM, ordinary pretraining, and offline teacher--reference corpus selection under matched training-time FLOPs that include online teacher computation~\citep{2410.17215}. MiniLLM receives fewer optimization steps and falls below ordinary pretraining on the downstream average across all three student sizes, whereas offline corpus selection improves that average. Teacher guidance can therefore be an effective substitute for one form of training while costing more than the benefit it supplies in another regime.

Whether this advantage survives over the full pipeline depends on how the teacher and its targets are obtained. Qwen3's stage comparison excludes construction of the stronger teacher; the sparse-to-dense reward pipeline explicitly spends compute improving a teacher before transfer~\citep{2605.12483}. MiniPLM reports offline scoring and reference-model preparation separately and considers amortization across students, so its matched training-time result does not settle the total-cost ordering. Lightning-OPD includes collection and precomputation in its totals, but obtains those savings with fixed states. The relevant comparison thus depends both on the alternative use of computation and on which preparation costs can be shared across later training.

Target representation also changes the storage and communication burden. AsyncOPD uses sparse and Monte Carlo estimators in its measured asynchronous pipeline~\citep{2606.24143}. DeepSeek-V4 instead performs full-vocabulary reverse-KL transfer from more than ten specialists, caching final hidden states for logit reconstruction and grouping samples by teacher so that one prediction head resides on device at a time~\citep{deepseekv4}. Asynchronous state and parameter transfers support this dense design, whose end-to-end speedup is not isolated in the report.

Reducing differentiation cost presents a different opportunity from reducing access to teacher predictions. SparseOPD constructs the full reverse-KL correction without retaining its gradient graph, then backpropagates through selected logits while compensating for omitted positive and negative correction mass~\citep{arxiv260934386}. Matched-support comparisons favor this correction over selecting only high-probability candidates. Memory savings are largest relative to the additional allocations above the pre-step baseline. The reduction in total peak memory is smaller, and complete updates can take longer. This makes sparsification useful for a memory bottleneck without establishing a training-speed advantage.

Trajectory repair couples teacher scoring to generation. Speculative Knowledge Distillation accepts student proposals in the teacher's top-$K$ set and resumes from a teacher replacement at the first rejection~\citep{2410.11325}. Its corrected trajectories require online verification and additional teacher sampling, unlike scoring a completed student response. The resulting cost depends on how frequently generation crosses between the two models.

Efficiency methods therefore trade computation against state coverage, freshness, and retained capability. Which of these changes explains a gain is a separate question from whether the complete method improves performance, motivating the mechanism studies in the next chapter.

%% file: figures/chapter_maps/08_training_budget.tex
\begin{figure}[!htb]
\centering
\includegraphics[width=\linewidth]{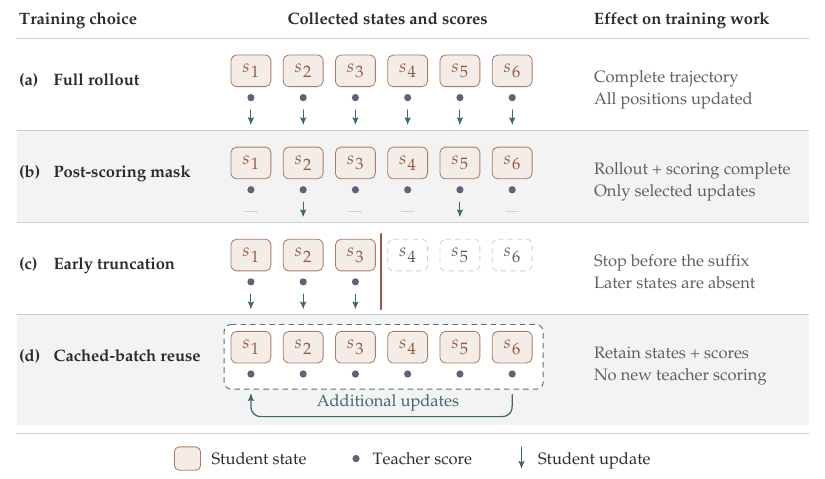}
\caption{Efficiency choices affect different work and training states. Post-scoring masks reduce supervised positions after generation and teacher scoring. Early stopping avoids the suffix and its states. Cached-batch reuse adds updates at stored prefixes without new collection or scoring. The trajectories and marker counts are schematic.}
\label{fig:training_budget}
\end{figure}

%% file: chapters/09_understanding.tex
\section{Understanding and Evaluating OPD}
\label{sec:understanding}

OPD gains can arise from teacher information, behavior elicited by privileged context, or update effects that persist without accurate teacher guidance. These explanations lead to different choices of supervision. They also bear on whether improved task accuracy comes with broader solution coverage, reliable stopping, and retained capabilities.

\subsection{Teacher Reliability and Student Learnability}
\label{subsec:teachability}

The reliability of teacher predictions and the student's ability to learn from them are distinct questions. Section~\ref{sec:signal} examines how to construct such guidance: continuation experiments and trajectory-refinement analyses distinguish the teacher's ability to solve a task from the reliability of its guidance at student prefixes~\citep{2604.13016,2605.30833,2606.08432}. Here the question is what explains transfer once that guidance is supplied. The evidence points to interactions among target representation, student initialization, and the update, rather than independent benefits from each component.

Compatibility depends on how the target is represented and fitted. Restricting vocabulary support can help, yet its interaction with special-token masking varies across tasks; an unnormalized top-$K$ reverse-KL implementation can also introduce a harmful gradient term~\citep{2603.25562,2605.11182}. A tighter variance bound for a token-local estimator does not remove its bias relative to sequence-level reverse KL~\citep{2603.25562}. These results explain why target restriction, masking, and estimator choice cannot be assumed to contribute independent gains.

The teacher--student pairing adds another dependence. Changing the student from base to post-trained initialization can reverse teacher rankings, and a log-ratio compression rule that helps one pairing can lose to hard clipping under a larger mismatch~\citep{2607.13399}. Choosing a teacher and regulating its influence are therefore coupled decisions.

Persistent disagreement need not produce a strong learning signal. An analysis of idealized full-KL OPD separates loss change into learning at fixed prefixes and a term induced by the changing rollout distribution~\citep{arxiv261011247}. Its local recovery bound applies to sufficiently nearby teachers sharing the student's parameterization; it does not guarantee recovery from an arbitrary stronger teacher. Training plateaus and declining gradient proxies are consistent with weak usable supervision, but the observed proxies neither isolate its cause nor measure the idealized flow directly.

Learnability also varies within a response. A shrinking teacher top-two log-probability gap on student candidates can identify prefixes beyond which additional supervision becomes less useful~\citep{2605.13643}. Coordinating token selection with the amount of privileged teacher information similarly improves allocation in the tested settings~\citep{2608.08176}. The first changes where supervision is applied; the second also changes the information behind it. These findings support adapting supervision to the student, but neither makes a distribution score a calibrated correctness estimate.

Prompt success rate has different implications for reward learning and teacher matching. Competence-based weighting has a gradient signal-to-noise motivation under explicit assumptions~\citep{2603.11178}. An all-wrong group provides no successful sibling for teacher conditioning. If its responses also receive equal rewards, their group-relative reward advantages are zero. A separate teacher-matching loss can nevertheless remain nonzero. Calibration and early length stability reveal further aspects of supervision that success rate alone misses~\citep{2604.16830,2604.08527}.

Figure~\ref{fig:mechanism_controls_map} connects competing explanations of OPD gains to the behaviors they leave unresolved.

\input{figures/chapter_maps/09_mechanism_controls}

\subsection{Transfer and Competing Explanations}
\label{subsec:mechanisms}

An accuracy gain does not by itself identify what the student has learned from the teacher. Three explanations recur in the evidence: acquisition of missing information, more effective execution of existing knowledge, and changes in the response distribution that do not require accurate teacher credit. These mechanisms can coexist. The comparisons below distinguish them by varying the objective, teacher context, or dependence of the update on teacher predictions.

\textbf{Information acquisition depends on the matching objective as well as the visited states.} Crossing teacher- and student-generated prefixes with forward and reverse KL separates two choices often changed together~\citep{2605.16826}. In synthetic lookup-and-composition tasks, forward KL improves factual transfer under either prefix source, while student prefixes improve the likelihood of correct multistep completions under either divergence~\citep{arxiv261009639}. Reverse-KL OPD more readily transfers execution to unseen composition trees than facts absent from the student's initialization; factual transfer is also limited in a natural-language recent-facts test. The crossed synthetic result nevertheless rules out interpreting this as a general inability of OPD to transfer knowledge. In the framework of Section~\ref{subsec:support}, changing the states exposes different learning situations, while changing the divergence reweights the alternatives at those states. The observed division between factual acquisition and execution is evidence for that distinction in these tasks, not a universal ranking of divergences.

The information supplied in context and the teacher's ability to interpret it are also distinct. DOPD compares a stronger teacher with a privileged-input view of the student under the same context, routing supervision using their absolute sampled-token log-probability gap together with their relative predicted probabilities~\citep{2606.30626}. Removing high-gap positions hurts more than removing an equal fraction of low-gap or random positions. This identifies useful guidance within the recipe, although calibration, style, and model interfaces prevent reading the gap as a pure measure of capacity.

Useful guidance need not locate the errors that matter for the final answer. Predictive discrepancy compares model probabilities, whereas correctness discrimination asks whether a score separates correct from incorrect responses. A small discrepancy does not imply a valid response, since most sampled tokens in an incorrect response can still receive similar probabilities from teacher and student~\citep{2607.23731}. At the response level, privileged-likelihood scores show near-chance discrimination and lower accuracy than an outcome-only control on AIME, but above-chance discrimination and improved held-out accuracy in biology~\citep{2608.09263}. The settings differ in both model and feedback format, so the contrast does not isolate a task effect. Limited AIME training replication and a context-template mismatch further constrain the comparison.

Locating an invalid reasoning step requires evidence beyond response-level discrimination. UECR compares its task credit with independently annotated invalid steps and finds an association that weakens when entropy positions are shuffled~\citep{2609.28385}. This tests whether the signal tracks where reasoning fails, rather than merely whether a response is correct. Even successful localization leaves a further learning question open, namely whether using that signal improves the student's subsequent reasoning.

\textbf{Privileged context can change execution without supplying the instance's answer.} A reverse-KL analysis distinguishes reusable latent rules from instance-specific solutions and shows that aggregating conditioned teachers produces a geometric target through their log probabilities~\citep{2605.11182}. Under forward KL, OP2SD often improves most when an unrelated worked mathematical example replaces the paired solution~\citep{2608.09228}. The paired answer is unnecessary for those gains, but trivial mathematical and physics contexts fail. This evidence supports a behavioral role for context while leaving its precise mechanism unresolved; it does not make context content irrelevant.

Improved execution must also be distinguished from repairing incorrect reasoning. Correct-only OPSD shortens thinking traces with student-dependent accuracy changes, whereas incorrect-only training can shorten responses while reducing accuracy~\citep{2605.06188}. Prompt matching and additional seeds preserve this ordering; richer context and longer training do not establish reliable repair. In complementary experiments, solution-conditioned self-teachers favor the supplied path, assign similar losses to correct and incorrect rollouts, and degrade explicit-thinking accuracy when distillation is the sole objective~\citep{2608.04794}. These results support compaction of existing reasoning more clearly than correction in the examined settings, leaving reward--distillation hybrids as a separate question.

Feedback-conditioned multi-turn RL can also improve first-turn performance without a distillation objective. Multi-turn GRPO does not reproduce the first-turn gain of RLTF-SD on Knights and Knaves, but already produces a large gain on Binary Matrix~\citep{2602.02482}. The additional benefit of distilling feedback-conditioned responses therefore depends on the task and training arrangement (Section~\ref{sec:rl_opd}). Access to feedback during interaction alone does not isolate this benefit.

\textbf{Some sampled-token gains persist when teacher-dependent credit is removed.} In one study, the sign of the teacher-derived advantage on verifiable final-answer tokens often disagrees with response correctness, yet trajectory subsets separated by this agreement achieve similar accuracy~\citep{2608.31046}. Supervising only the lowest-probability fifth of student tokens reproduces the tested OPD result. With that same mask, a fixed negative coefficient performs comparably to teacher advantages, whereas a fixed positive coefficient collapses training. An entropy-modulated teacher-free variant improves further while retaining on-policy sampling. Suppression of low-probability tokens sampled by the student can therefore explain the gain in this implementation. The result concerns a particular sampled-token loss; it does not establish equivalence for full-distribution matching, other estimators, or substantially larger students.

The practical consequence is that a successful recipe need not support every intended form of transfer. Compaction of successful reasoning does not imply error repair, just as reproducing one loss without teacher credit does not make teacher information dispensable elsewhere. Table~\ref{tab:mechanism_controls} summarizes which explanations the interventions distinguish. Their differing objectives and students support several mechanisms, rather than a single account of all OPD gains.

\begin{table}[!htbp]
\centering
\small
\setlength{\tabcolsep}{4pt}
\renewcommand{\arraystretch}{1.08}
\begin{tabularx}{\linewidth}{@{}P{0.30\linewidth}P{0.31\linewidth}>{\raggedright\arraybackslash}X@{}}
\toprule
\textbf{Intervention} & \textbf{Supported conclusion} & \textbf{Remaining uncertainty} \\
\midrule
Cross teacher/student prefixes with forward/reverse KL~\citep{2605.16826} & Prefix source and conditional divergence are separate choices. & Changing both together cannot identify the contribution of either. \\
\addlinespace
Replace the paired solution with an unrelated worked example~\citep{2608.09228} & Target-specific answers are unnecessary for the reported forward-KL gains. & Useful contexts still have structure and content; one training run per condition limits certainty. \\
\addlinespace
Train on correct or incorrect rollouts, including a same-prompt control~\citep{2605.06188} & The tested reasoning recipe supports compaction more clearly than repair. & Accuracy costs depend on the student; reward--distillation hybrids are not tested by this control. \\
\addlinespace
Replace teacher advantages on the lowest-probability fifth of student tokens with a fixed negative coefficient~\citep{2608.31046} & Teacher-dependent credit is unnecessary to reproduce the tested sampled-token baseline. & The result does not establish equivalence for full-distribution matching or larger students. \\
\bottomrule
\end{tabularx}
\caption{Mechanism controls and the scope of their conclusions. The interventions probe different alternatives to a general knowledge-transfer explanation.}\label{tab:mechanism_controls}
\end{table}

\subsection{Failure Modes and Preservation Targets}
\label{subsec:geometry}

The mechanism studies above limit what can be inferred from an accuracy gain. Preservation raises the complementary question of what can be lost: alternative successful paths, usable output behavior, or properties that routine task scores do not measure.

\subsubsection{Distributional concentration and solution coverage}

Finite-budget head--tail analyses qualify simple mode-covering and mode-seeking descriptions~\citep{2404.02657}. In matched mathematical comparisons, reverse KL improves Avg@k over forward KL by $2.45$ points on average, yet the 4,096-token setting also exhibits low predictive entropy and severe response-length inflation~\citep{2605.16826}. Conversely, the failing runs with self-distillation as the sole objective discussed above flatten the student distribution~\citep{2608.04794}. Degradation can therefore accompany opposite changes in entropy.

Controlled entropy comparisons find a stronger effect of divergence choice than of teacher- versus student-prefix sampling, with teacher refresh and tail treatment also affecting the outcome~\citep{arxiv261003529}. A corresponding forward-KL analysis relates average student and teacher entropy at a stationary solution when the student family is closed under logit scaling and both entropies are measured on the same contexts. This is not a tokenwise or training-step guarantee, and reverse KL can also increase entropy in some regimes. Entropy describes distributional concentration; its direction alone does not establish whether task ability is preserved.

Target temperature can change the relative performance of divergences. Softening teacher and student distributions reverses the FKL--RKL ranking in one study of fixed-response instruction distillation~\citep{2606.00306}. This reversal is observed in \emph{off-policy training on fixed teacher-generated responses}; its extension to OPD remains open. Temperature changes teacher targets for FKL and both the scale and student-probability weighting of the RKL gradient. With fixed centered logits, their high-temperature gradients share a leading logit-matching term, although their finite-temperature dynamics can differ.

\textbf{Success on one attempt and coverage across repeated attempts can move differently.} A sequence-level analysis shows how conditioning on correct demonstrations can amplify differences between equally correct paths~\citep{2606.26091}. Its optimum reweights paths by the average change in their log probability under demonstration conditioning. This result assumes a frozen base-policy teacher and a fixed distribution of correct demonstrations, and averages divergences to conditioned targets rather than matching their arithmetic mixture.

Empirically, in the same study, competitive pass@1 coexists with lower pass@16 than GRPO, and multiple correct demonstrations do not fully restore solution coverage~\citep{2606.26091}. Separate graph experiments likewise find that diverse external demonstrations do not recover semantic diversity. These observations are consistent with concentration reducing coverage, without establishing that the frozen-teacher mechanism fully explains the training results. Likelihood-based precision--coverage measurements in simulations and offline pretraining quantify yet another property and cannot be equated with correctness-based pass@$k$~\citep{2505.13111}.

Coverage loss is not universal across OPD settings. A test-time scaling study finds that pass@$k$ gains can persist or grow at larger sampling budgets in some comparisons and reverse in others~\citep{arxiv260811829}. Increasing the budget changes how per-problem success probabilities contribute to aggregate pass@$k$, so mixed improvements and regressions can change the method ranking. A local full-vocabulary reverse-KL analysis explains how such reversals are possible; it does not predict the finite-step training dynamics of every implementation. Evaluating multiple sampling budgets therefore reveals improvements and regressions that a single pass@1 average can conceal.

\subsubsection{Reasoning behavior, confidence, and usable outputs}

Solution coverage measures which problems can be solved across attempts; changes in the form of those attempts require separate interpretation. Analyses of limited marker sets and annotated trajectories associate reasoning degradation with suppression of uncertainty expressions. The solution-conditioned teachers examined in these analyses also lower probabilities of several deliberative tokens~\citep{2603.24472,2605.11609}.

Thinking-collapse interventions make the distinction between a behavioral marker and task ability concrete~\citep{2607.10805}. Masking high-entropy positions preserves more of the measured thinking markers than ordinary self-distillation, yet does not recover downstream accuracy. AD-OPSD instead mixes the privileged target with a frozen base-model reference using a local disagreement gate. Its comparisons support reference mixing in the tested settings, while the additional benefit of dynamic gating varies with the student. Restoring marker frequency alone therefore remains an insufficient test of reasoning preservation.

CaOPD addresses the confidence mismatch between a teacher that sees privileged information and a deployed student that does not~\citep{2604.16830}. It replaces verbal-confidence values in both scored completions and teacher context with empirical student success rates, improving reported calibration while retaining accuracy. The method requires identifiable confidence tokens and sampled student outcomes, which can reuse existing SDPO rollouts. Its target is confidence that tracks correctness, a property distinct from the frequency of uncertainty expressions or predictive entropy.

Usable behavior can depend on small distributional differences that aggregate matching obscures. Stable-OPD mitigates length instability through reference anchoring and off-policy supervision, whereas a termination study identifies a specific failure to transfer probability between functionally equivalent stop tokens~\citep{2604.08527,2609.20511}. Probability-level alignment repairs stopping where simply expanding the decoding stop set fails, but does not consistently improve accuracy. A multi-teacher study finds a related problem for tool use: top-$K$ truncation can omit a decision-critical tool-entry token despite retaining almost all teacher probability mass~\citep{2607.07050}. Restoring its supervision reduces erroneous calls at a cost to required calls. Functional equivalence and decision relevance therefore matter alongside the amount of probability mass matched.

Privileged-context leakage and format failure can hide beneath aggregate scores. Leakage occurs when the student refers to information available only in the teacher's context, leaving the reader unable to resolve the reference. Such references increase across full-distribution, teacher-top-1, and student-top-1 supervision variants as later validation accuracy deteriorates~\citep{2604.03128}. ErrOPD shows that extrapolation beyond the teacher can sharply reduce parsing validity when errors have already transferred or are carried by frequent student decisions~\citep{2605.08737}. Error transfer also depends on training duration, and ranking metrics restricted to successfully parsed responses can mask the resulting failures. Quality among usable outputs can therefore conceal a deterioration in their availability.

Successful internalization need not remain robust when the privileged context returns. No-Context Anchoring penalizes disagreement between predictions of a shared model with and without additional privileged context, using the branch without that context as a stop-gradient target~\citep{2606.11627}. It reduces context-induced harm in twelve of fourteen evaluated configurations. The benefit comes with a tradeoff when internalization is incomplete and the reintroduced context still supplies useful information.

Even faithful matching can optimize an imperfect proxy for the task. At fixed prefixes, reverse-KL minimization maximizes teacher log probability plus student entropy; repetitive or overly long outputs can become more likely while that objective improves~\citep{arxiv261003185}. A separate study observes improving teacher-proxy scores alongside worsening alignment with an independent oracle under repeated offline reuse~\citep{2502.02671}. Refreshing responses from either teacher or student mitigates the latter effect, supporting data renewal more directly than a student-only sampling explanation. These failures distinguish poor optimization of a useful target from successful optimization of a target that does not fully capture desired behavior.

\subsubsection{Hidden behavior and training-data exposure}

Teacher matching can also transfer behaviors that ordinary safety evaluations miss. With a deliberately backdoored teacher and trigger-bearing distillation prompts, an initially clean student learns trigger-dependent harmful behavior while improving refusal on trigger-free harmful prompts~\citep{arxiv261007654}. The evaluated attack combines a compromised teacher with a poisoned prompt distribution; it is not evidence of spontaneous backdoor formation under clean supervision. Log-ratio clipping delays transfer in some tested Qwen2.5 settings but does not reliably prevent it as training continues.

Training membership presents a different privacy question: can a change in teacher agreement reveal which prompts participated in learning? Teacher Alignment Gain measures the reduction in teacher discrepancy between the pre- and post-distillation students on the same contexts generated by the pre-distillation student~\citep{arxiv261011423}. Its membership-inference comparisons find information beyond the amount of student-distribution drift alone. This privacy exposure is demonstrated under access to both students' full next-token distributions, a teacher or compatible surrogate, and labeled member and nonmember references from the same training run. It therefore concerns a richer access setting than querying the deployed student's responses alone.

\subsubsection{What parameter changes reveal about preservation}

Parameter geometry offers a way to ask whether successful adaptation requires widespread change. Its usefulness for training depends on when the relevant structure becomes available. Rescaling intermediate module updates to their eventual norms recovers much of the final performance in the study introducing EffOPD, but requires future information~\citep{2605.11739}. EffOPD turns this diagnostic into a usable intervention by extrapolating recent parameter displacements and validating the result, improving the reported accuracy--time tradeoff with validation overhead included. The distinction is between explaining a completed trajectory and exploiting its structure while learning is still underway.

The same distinction applies to low-dimensional and sparse updates. Projecting gradients onto a small set of directions estimated during training preserves much of the measured OPD performance, whereas effective sparse retraining can use parameter coordinates selected after a full training run~\citep{2606.07082,2606.13657}. The former estimates a useful subspace online; the latter reproduces adaptation using a retrospectively identified mask. These structures are also mathematically distinct: concentrated update magnitude can coexist with full numerical rank, and controls retaining similar rank dynamics do not identify on-policy sampling as their cause. Both lines of work clarify how cheaply adaptation might be represented, but compact parameter changes alone do not establish which prior capabilities survive.

\subsection{Generalization and Subsequent Learning}
\label{subsec:evaluation}

The value of learning on student states depends on whether useful guidance remains available beyond familiar conditions. Theory clarifies one part of this dependence: interactive teacher access can improve sample complexity under structured noise. For a finite policy class containing the clean expert and a known corruption process, offline imitation can require exponentially many samples in the decision horizon, whereas interactive access permits polynomial dependence~\citep{2606.30923}. Horizon-independent bounds additionally restrict corruption over actions supported by the expert and candidate policies. With unknown corruption, NAILGUN requires a deterministic expert, bounded per-action corruption, and a margin separating expert behavior from noise. Interactive access is therefore valuable under conditions that make the clean behavior identifiable; standard OPD still fails at high noise in the reported experiments.

Synthetic planning comparisons examine the favorable case of learning under oracle supervision: the teacher's described training source includes evaluation instantiations~\citep{2607.24720}. They illuminate how a student can learn when guidance is available, while leaving its availability in unseen environments unresolved. Equal-compute advantages and transfer to natural environments likewise require evidence beyond this oracle setting.

Even when guidance improves the training task, the resulting policy may transfer poorly to other domains. A vision-language OPD baseline improves its training modality slightly while losing video and pooled language--video performance~\citep{arxiv260618216}. Differing correctness filtering prevents attributing the loss solely to prefix source, but the recipe exposes a retention cost concealed by the training-modality gain. Validation can also favor the wrong transfer behavior: in another vision-language study, method rankings reverse their relationship to cross-domain performance when the validation slice changes~\citep{2608.21811}. Although the cause is not isolated, a convenient within-domain score can select a different recipe from the one that transfers best.

Capability integration and subsequent RL reveal further differences concealed by intermediate averages. Under matched expert pools, small average differences among fusion methods coexist with substantial benchmark-specific differences~\citep{2608.27409}, as discussed in Section~\ref{subsec:mopd_allocation}. The study in Section~\ref{subsec:sequential_training} similarly observes little immediate OPD improvement but a larger advantage after shared downstream RL, alongside a change in the preferred divergence~\citep{2609.28145}. A method's immediate score can therefore differ from its value as preparation for later learning. These outcomes concern retained capability and trainability, while the full training costs remain a separate comparison (Section~\ref{subsec:systems}).

Teacher selection inherits the same distinction. NPD compares two teachers reached along one GRPO trajectory: the nearer teacher gives a lower immediate validation score but a higher average across eight benchmarks after common downstream RL~\citep{arxiv260420733}. Another study finds that teacher stages along instruction-following RL trade off instruction transfer against mathematical retention, with the tested converged teacher failing to preserve mathematics~\citep{arxiv260800220}. These results motivate selecting teachers for the student's intended continuation and retained abilities, rather than teacher accuracy alone. Different distillation durations and single-run comparisons leave the optimal teacher stage unresolved; neither study isolates teacher distance as a universal cause of later learning success.

%% file: figures/chapter_maps/09_mechanism_controls.tex
\begin{figure}[!htb]
\centering
\includegraphics[width=\linewidth]{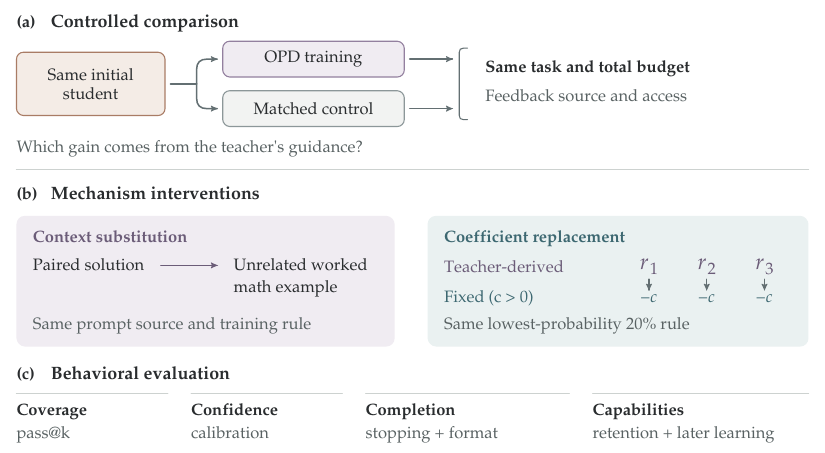}
\caption{From observed gains to tested explanations. (a) Matched comparisons distinguish the contribution of guidance from other training differences. (b) Two separate studies replace paired solutions with unrelated context~\citep{2608.09228}, or teacher-dependent coefficients with a fixed negative constant while retaining the lowest-probability token-selection rule~\citep{2608.31046}. The rules remain fixed while trajectories evolve. (c) Behavioral measures include coverage, calibration, completion, and retention.}
\label{fig:mechanism_controls_map}
\end{figure}

%% file: chapters/10_applications.tex
\section{Applications and Practical Guidance}
\label{sec:applications}

OPD serves several deployment goals, from transferring capability to a smaller model and consolidating specialists to changing a model's generation process or numerical precision. Each places a different demand on supervision. Compression emphasizes capability retention, consolidation balances specialist strengths, and model conversion exposes new states or constraints that the teacher must address.

\subsection{Capability Transfer}
\label{subsec:deployment}

Capability transfer first requires deciding whether teacher-generated examples suffice or supervision must follow the student's own behavior. Large-model pipelines illustrate both choices. DeepSeek-R1 obtains strong smaller models through curated teacher traces, while Qwen3 follows supervised initialization with teacher logit supervision on student-generated responses~\citep{2501.12948,2505.09388}. The practical distinction is between supplying useful examples and correcting behavior encountered during student generation. Qwen3's favorable comparison with direct RL concerns the student stage and excludes teacher preparation.

Consolidation adds the requirement to preserve several useful behaviors in one policy. MOPD and Open-MOPD address allocation and optimization imbalances that can leave specialist skills underrepresented~\citep{2606.30406,2608.19098}. ORBIT applies a related idea to reasoning budgets, fusing experts trained under different context limits into budget-conditioned modes~\citep{2601.08310}. Its shortest mode loses substantial accuracy relative to longer modes, while longer modes provide different accuracy--length choices that prompt steering alone does not reproduce. Domain consolidation seeks complementary skills; budget consolidation seeks controllable operating points. Both require evaluating the retained choices individually, with expert preparation included in the training cost.

Acquisition must also be weighed against retention. MixSD varies the mixture of fact-conditioned and original-input targets, exposing cases where preserving existing capability costs target-task performance~\citep{2605.16865}. Its fixed-base target generation provides an alternative to repeated token-level matching on current-student rollouts. Safety adaptation poses a similar tension: privileged safety and helpfulness guidance improves composite scores in OPSA, but adaptive-attack results remain mixed~\citep{2605.15239}. Robustness to forced unsafe prefixes improves against the examined comparator while iterative attacks expose regressions in several models. Distillation can therefore transfer a desired behavior without uniformly preserving adjacent capabilities or robustness.

Figure~\ref{fig:deployment_interfaces} summarizes the generation, precision, and action interfaces through which OPD changes deployed behavior.

\input{figures/chapter_maps/10_deployment_interfaces}

\subsection{Deployment and Model Conversion}
\label{subsec:inference}

Acceleration depends first on who is responsible for the final prediction. With a verifier at deployment, a draft model aims to propose tokens the larger target will accept. DistillSpec aligns the draft distribution for this purpose; AdaFlash continues adapting a diffusion drafter from verification feedback during serving~\citep{2310.08461,arxiv260719223}. Acceptance alone does not determine speed: draft generation, verification, scheduling, and adaptation all consume resources. AdaFlash uses additional training GPUs, and its speedup can fall below one at high concurrency in an evaluated configuration.

Without a verifier, the converted student's predictions must stand on their own. MTP-SD trains a standalone multi-token predictor on student-generated spans~\citep{2602.06019}. Its model-dependent accuracy loss and smaller acceleration gains on open-ended generation tasks expose a different tradeoff from draft acceptance. Effective progress above three tokens per step in selected settings is distinct from the serving-throughput gains measured separately in SGLang. The deployment contract thus changes both the supervision target and the meaning of acceleration.

Changing the generation process also changes the states requiring supervision. A diffusion student refines partly specified sequences rather than extending left-to-right prefixes. TABOM supplies local denoising-order supervision on reused base-model trajectories, while autoregressive-to-diffusion distillation follows the diffusion student's own denoising trajectories~\citep{2605.11854,2606.06712}. The first addresses ordering within a local objective; the second aligns guidance with the converted student's evolving states. Local gains do not establish equivalence to global distribution matching, and conversion-stage token counts alone do not establish total-compute savings.

Architectural and numerical conversion introduce a further requirement: the student must remain viable enough to learn from the teacher. In SpikeOPD, ordinary full-vocabulary forward-KL OPD degenerates into repetition, motivating teacher correction, reference anchoring, and firing-rate controls~\citep{2608.27857}. Quantization-aware OPD instead supervises behavior generated through the low-bit forward path, combining teacher feedback with verifier optimization to recover accuracy~\citep{2609.26708}. Both seek stable guidance after a change in architecture or precision. SpikeOPD addresses student-prefix mismatch and drift in spiking dynamics, with small-scale ablations supporting its three components; the low-bit results build on a viable quantization-aware-distillation initialization. Analytical energy estimates for spiking models and accuracy recovery for quantized models also leave measured hardware speedup as a separate question.

\subsection{Structured and Multimodal Use}
\label{subsec:domains}

When the output is a spatial prediction, action, or formal artifact, the supervised object may differ from the deployed decision. A language judgment can guide a visual prediction, a hidden representation can support an answer, and an action annotation can guide interaction. Transfer depends both on how supervision reaches that decision and on which aspects of success the feedback covers.

For visual tasks, a language-based teacher must supply guidance that changes the modality-specific decision. Selective text-teacher supervision contributes to visual reasoning in VOLD, but its additional benefit depends on the preparation data used alongside reward updates~\citep{2510.23497}. MOPDA makes the connection more explicit by using a distilled language judgment to contextualize visual anchors for industrial anomaly localization; its EMA teacher receives verified defect analysis~\citep{arxiv260718850}. Disrupting that contextualization or substituting incorrect judgments weakens localization, showing how the intermediate judgment contributes to the spatial decision. The gain concerns localization and does not establish superiority on every image-level metric.

Intermediate language is not always necessary. After recurrent latent computation, Lapras aligns the student's hidden states used to predict the first answer token with those of a teacher conditioned on verbal reasoning~\citep{arxiv261011111}. This supports time-series question answering without intermediate token generation. Removing distillation while retaining latent computation lowers accuracy in the tested configuration, although the smallest backbone remains below direct answering in average F1. MOPDA and Lapras thus expose different interfaces for transferring reasoning: a language judgment that affects a spatial decision, or a latent representation that supports the answer directly. The useful interface depends on the task and the student's capacity.

In graphical user interface (GUI) tasks, privileged visual hints or human-verified action annotations provide guidance on student-generated coordinate or action predictions~\citep{2605.00642,2605.07505}. The challenge is to connect this local supervision to successful interaction, especially when several actions are valid. LiteGUI matches each student output to an annotated valid action for teacher conditioning, then refines the policy with action-matching and subtask-planning rewards. This accommodates alternatives within the annotations, but a finite action set can still penalize valid choices outside it. Agreement with an annotated action can therefore diverge from successful task completion.

Structured-output transfer depends on \emph{what the verifier can check}. RWOPD weights supervision for generated SystemVerilog Assertions using a property-equivalence checker~\citep{2605.13501}. The checker abstains on liveness properties, which concern whether an event eventually occurs; gains there consequently fall outside direct verifier-weighted supervision. Preparation and filtering also contribute to the recipe. A formal verifier can strengthen supervision within its coverage without certifying the entire task.

These cases distinguish a useful supervision interface from complete task feedback. Language judgments and latent representations can influence a downstream decision without reproducing the teacher's intermediate reasoning. Action annotations and formal verification can constrain that decision, but only within the choices or properties they cover. Successful transfer therefore requires both a path from supervision to behavior and an account of what remains outside the feedback's reach.

\subsection{Matching OPD to Deployment Needs}
\label{subsec:guidance}
\textbf{The useful OPD design depends on which supervision is available and which behavior deployment requires.} Teacher--student size ratios alone do not determine that choice. Table~\ref{tab:practical_guidance} relates deployment needs to the conditions under which the preceding methods apply and the alternatives that clarify their value.

Teacher access is an initial constraint. Matching explicit teacher probabilities requires scores for comparable events; text-only access instead supports continuations, critique-conditioned targets, or response-level rewards, with different information and cost (Section~\ref{subsec:teacher_roles}). A verified outcome can guide selection or reward refinement, but it cannot supply missing token probabilities. Conversely, teacher probabilities can support matching where no verifier exists, without establishing that agreement improves task success. Interface alignment and the coverage of task feedback therefore determine which combination is meaningful.

\begin{table}[!ht]
\centering
\small
\setlength{\tabcolsep}{3pt}
\renewcommand{\arraystretch}{1.08}
\begin{tabularx}{\linewidth}{@{}P{0.18\linewidth}P{0.26\linewidth}P{0.25\linewidth}>{\raggedright\arraybackslash}X@{}}
\toprule
\textbf{Deployment need} & \textbf{Condition for use} & \textbf{Candidate design} & \textbf{Comparison that matters} \\
\midrule
Improve a student & Initial rollouts admit useful teacher guidance; explicit probability matching has compatible score access & Supervised preparation if needed, then supervision on student states & Offline continuation from the same initialization; direct RL when rewards are available \\
\addlinespace
Retain specialist skills & Experts offer complementary capabilities and each domain receives adequate supervision & Routed multi-teacher transfer with allocation across domains & Merging or pooled RL; per-domain retention and total expert preparation cost \\
\addlinespace
Meet several reasoning budgets & Budget-specific experts provide useful accuracy--length choices & Budget-conditioned expert transfer & Prompt-only control at comparable output lengths; accuracy at each budget \\
\addlinespace
Convert the model or decoder & The converted student is viable and teacher guidance can be aligned to its decisions & Draft alignment or supervision on converted-student states & Unrefined or offline-refined conversion; task quality and measured latency \\
\addlinespace
Internalize privileged feedback & Extra context improves guidance at student states; reward refinement also needs a suitable reward signal & Conditioned-teacher supervision, with reward refinement where available & Direct use of the same feedback or annotations; reward-only learning with the same signal \\
\bottomrule
\end{tabularx}
\caption{Conditional guidance for selecting an OPD design. The conditions describe the demands of each use case, not sufficient guarantees of improvement. Supporting comparisons appear in Sections~\ref{subsec:deployment}--\ref{subsec:domains}.}\label{tab:practical_guidance}
\end{table}

The dominant cost determines which efficiency choices are relevant. Expensive generation favors horizon control when omitted states do not require direct supervision. Expensive teacher scoring favors caching or batch reuse when the acquired targets remain useful, subject to the freshness tradeoffs in Section~\ref{subsec:freshness}. A memory bottleneck may favor sparse differentiation even without faster updates. Each choice must address the limiting resource under the full cost accounting in Section~\ref{subsec:systems}.

The intended learning sequence also changes the choice of objective. For immediate deployment, a matching objective should support the required behavior, including solution coverage, reliable termination, and retained capabilities. For a student that will undergo further RL, its value depends on performance after that continuation. The forward-/reverse-KL reversal in Section~\ref{subsec:sequential_training} makes immediate accuracy an insufficient basis for choosing between them. Available supervision determines which updates are possible, resource constraints determine which are practical, and the intended use determines which outcomes matter.
\FloatBarrier

%% file: figures/chapter_maps/10_deployment_interfaces.tex
\begin{figure}[!htb]
\centering
\includegraphics[width=\linewidth]{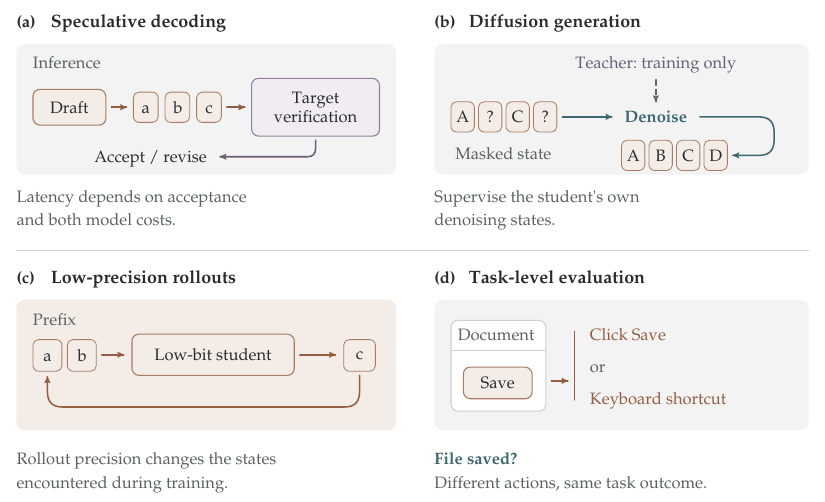}
\caption{Deployment interfaces change the transfer problem. (a) Speculative decoding retains target-model verification at inference, with latency depending on acceptance and both model costs. (b) A diffusion student receives training supervision on its denoising states. (c) Low-precision rollouts change the student's training distribution; inference speed also depends on deployed kernels. (d) Clicking Save and using a shortcut illustrate different actions with the same task outcome.}
\label{fig:deployment_interfaces}
\end{figure}

%% file: chapters/11_open_problems.tex
\section{Open Challenges and Future Directions}
\label{sec:future}

The central unresolved problem is how to adapt supervision to a changing student. Useful guidance depends on what the student has already learned, which states it will encounter next, and which capabilities must survive later training. This connects the prediction of learning benefit to compute allocation, persistent adaptation, and transfer across model interfaces.

Figure~\ref{fig:open_question_map} connects these open questions to the behavior each seeks to explain or preserve.

\input{figures/chapter_maps/11_open_questions}

\subsection{Downstream Value of Supervision}
\label{subsec:future_reliability}

Student uncertainty, teacher confidence, and teacher--student disagreement describe current predictions, whereas the value of supervision depends on how an update changes later behavior. Token-importance selection and prompt-level reliability gating address parts of this gap through distribution scores and teacher-rollout probes~\citep{2604.14084,2609.02998}. Yet privileged information can yield unreliable token credit, and extra context can change behavior without supplying the current problem's solution~\citep{2608.09263,2608.09228}. A predictor of learning benefit must therefore account for how information enters the update, rather than treating confidence or access to an answer as sufficient evidence of value.

More direct evidence of benefit is available, but its predictive reach and cost remain in tension. Meta-learned weighting measures the effect of a one-step update on reference solutions at substantial training cost (Section~\ref{subsec:selection}). FTB probes behavior after a teacher intervention through a teacher-preference proxy on paired student continuations~\citep{2608.01953}. It limits probing to one intervention per trajectory and short continuations, with faster training steps than vanilla OPD in the reported ALFWorld comparison. The contrast is between measuring a local learning effect and testing an immediate behavioral consequence. Neither yet provides an inexpensive predictor of longer-term benefit across students and training stages. Extending either approach must improve that prediction enough to justify computation that could otherwise be spent on learning.

The unresolved step is to connect affordable local evidence to the changing states reached through learning. An update may help at the current prefix while moving the student toward histories where the same teacher becomes less useful. Predicting this effect requires separating the contributions of context and update design identified in Section~\ref{subsec:mechanisms}, while extending beyond a single update or continuation. This would make selection responsive to what guidance enables the student to learn next.

\subsection{Joint Allocation of Prompts, Rollouts, Teacher Queries, and Updates}
\label{subsec:future_scaling}

Fresh experience and repeated learning from existing states compete for the same training budget. The scaling relationships in Section~\ref{subsec:distillation_scaling} connect teacher quality and scoring cost to the resources available for student training. OPD couples these tradeoffs to prompt acquisition, student generation, refresh, and potentially environment interaction. The unresolved choice is which use of computation produces the greatest learning benefit at the student's current stage.

The difficulty is that the return from one resource depends on how the others are used. The value of additional prompts can depend on whether trajectories are renewed; few-prompt and data-free results likewise show why prompt count alone does not measure the training experience~\citep{2609.25048,2609.04172,2609.14193}. Shorter rollouts and cached targets save computation by giving up different forms of coverage or freshness~\citep{2602.15260,2604.13010}. In MOPD, allocating another teacher query also depends on what other specialists have transferred and how much token-level optimization each domain receives~\citep{2608.19098}. Component savings consequently cannot be combined without accounting for their effects on the experience and supervision available elsewhere in training.

An allocation rule would need to capture these interactions as learning proceeds. Fresh states can broaden experience, another teacher query can improve guidance on an existing state, and another update can extract more from guidance already obtained. Their marginal values are not yet characterized jointly, especially when the budget also includes preparing several specialist teachers. This leaves a gap between reported component savings and an adaptive allocation of total computation.

\subsection{Lifecycle Control under Persistent Adaptation and Interaction}
\label{subsec:future_adaptation}

Over a long learning sequence, supervision can become redundant or need to return. Demonstration-conditioned self-distillation supplies evidence for accumulating skills across successive tasks~\citep{2601.19897}, but does not yet explain how to anticipate interference before a capability deteriorates. The scheduling problem is therefore preventive as well as corrective: continued teacher access and rehearsal incur a cost, yet withdrawing them too early may create a later recovery problem.

The reference for deciding when to intervene can itself change. When a self-teacher shares the evolving student's parameters, reward updates can alter both its competence and the behavior that needs correction. In interactive settings, changes in tools, memory policies, or world-model errors also change the states on which this guidance is used. Adaptive horizons already connect supervision to visited interaction states~\citep{2607.05804}; extending this control across repeated episodes requires relating episode depth and exploration cost to what the student retains between episodes.

A schedule must also anticipate the next training stage. Joint and sequential recipes can produce different outcomes, and subsequent RL can change the preferred distillation objective~\citep{2609.04108,2609.28145}. Preserving current performance is thus only part of the problem: a policy must remain able to benefit from later reward learning and interaction. Connecting early signs of interference to later trainability would help determine when to retain, renew, or withdraw supervision without merely shifting a capability loss to the next stage.

\subsection{Behavioral Equivalence across Tokenizers, Representations, Modalities, and Architectures}
\label{subsec:future_interfaces}

Different interfaces create two related obstacles to transfer: models may express predictions in different units, or make those predictions from different information and decision states. Tokenizer mappings address the first by aligning textual units~\citep{2607.22334}. Cross-lingual safety transfer also faces the second when the teacher receives translated input and additional instructions unavailable to the student~\citep{2605.02971}. An aligned output space does not remove this information gap; the student must learn behavior that remains supported by its own input.

Changes in generation architecture make state alignment equally consequential. In diffusion language models, dOPSD supervises predictions at partially masked states using later states of the same trajectory, whereas trace-based TOPD matches intermediate decisions retained in the final response~\citep{2607.04428,2607.16872}. The token position alone does not identify the prediction problem, because the visible context changes during denoising. These constructions raise the broader question of when guidance across states remains valid and when teacher and student must be compared at the same decision state.

Ultimately, the mapping must preserve distinctions that matter for behavior. Internal alignment can improve without improving task performance~\citep{2609.28845}, and token or feature agreement may omit differences needed for actions, solution diversity, or safety. Characterizing which distinctions survive a representation change would clarify when distribution matching is an adequate target and when supervision must operate more directly on actions or outcomes.

%% file: figures/chapter_maps/11_open_questions.tex
\begin{figure}[!htb]
\centering
\includegraphics[width=\linewidth]{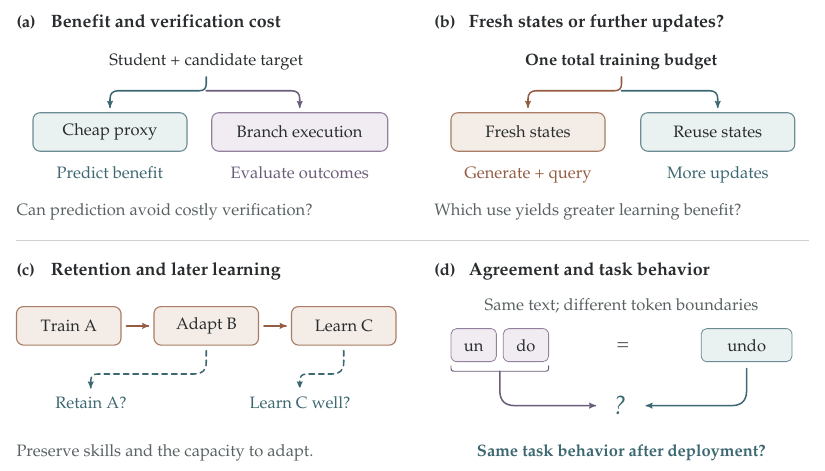}
\caption{Unresolved tradeoffs in adaptive supervision. (a) Can a target's learning benefit be predicted without paying the cost of verification? (b) When does generating fresh states offer more benefit than further updates on existing states under the same total budget? (c) Can adaptation preserve earlier capabilities and the capacity to learn later tasks? Dashed arrows denote evaluations. (d) When does agreement across prediction interfaces preserve task behavior? The panels are schematic.}
\label{fig:open_question_map}
\end{figure}

%% file: chapters/12_conclusion.tex
\par
\pagebreak[0]
\begin{samepage}
\section{Conclusion}
\label{sec:conclusion}

On-policy distillation connects teacher guidance to the experience generated by a learning student. This survey has organized that connection around states, target information, and policy updates, providing a common basis for understanding methods across different supervisors and training settings. The value of guidance depends jointly on what it conveys at student histories, how the chosen update uses it, and which states the resulting policy reaches. Choices about supervision and computation consequently shape the learning process together. Neither teacher strength nor a divergence considered in isolation explains when OPD will be effective.

\nopagebreak[4]
This perspective also changes what counts as successful transfer. Improved agreement or immediate accuracy may reflect different mechanisms and can coexist with lost solution coverage, weaker retained capabilities, or reduced benefit from later training. The value of OPD therefore lies in the behavior it makes available over the intended learning and deployment sequence, relative to the computation required. The challenge ahead is to adapt guidance as that sequence unfolds: to recognize when existing supervision remains useful, when new experience or a different supervisor is needed, and when continued imitation constrains further learning. Resolving this challenge would make OPD a more dependable means of transferring capabilities while preserving the student's capacity to develop beyond its current supervision.
\par
\end{samepage}